\documentclass[a4paper,fleqn]{cas-dc}

\usepackage[authoryear]{natbib}
\usepackage{float}
\usepackage{booktabs}
\usepackage{graphicx}
\usepackage{booktabs, amssymb, adjustbox}

\usepackage{makecell}

\usepackage{pgfplots}

\usepackage{placeins}
\pgfplotsset{compat=1.18}
\pgfplotsset{colormap={bluered}{rgb(0cm)=(0,0,1); rgb(1cm)=(1,0,0)}}

\usepackage{tikz}
\usepackage[table]{xcolor}

\def\tsc#1{\csdef{#1}{\textsc{\lowercase{#1}}\xspace}}
\tsc{WGM}
\tsc{QE}

\begin{document}
\let\WriteBookmarks\relax
\def\floatpagepagefraction{1}
\def\textpagefraction{.001}
\let\printorcid\relax

\shorttitle{}    

\shortauthors{}  

\title [mode = title]{SemLite-GS: Semantic-Selective Parallel Gaussian Splatting for Lightweight Building Reconstruction from UAV Oblique Imagery}  

\author[1]{Tengfei Wang\textsuperscript{\ensuremath{\dagger}}}
\author[1]{Xin Wang\textsuperscript{\ensuremath{\dagger}}}

\author[1]{Zihan Li}
\author[1]{Quanjian Ji}
\author[1]{Yifei Yu}

\author[2]{Dong Lin}

\author[1]{Zongqian Zhan}
\cormark[1]

\nonumnote{\textsuperscript{\ensuremath{\dagger}} Tengfei Wang and Xin Wang contributed equally to this work.}
\cortext[1]{Corresponding author: Zongqian Zhan.}

\affiliation[1]{
    organization={School of Geodesy and Geomatics, Wuhan University},
    city={Wuhan},
    postcode={430079},
    country={China}
}

\affiliation[2]{
    organization={Space Engineering University},
    city={Beijing},
    postcode={102299},
    country={China}
}


\begin{abstract}
Lightweight 3D building models are crucial for urban spatial analysis, yet existing pipelines typically involve costly dense scene reconstruction, building extraction, and mesh simplification. While 3DGS reconstructs scenes more efficiently than conventional multi-view stereo (MVS) pipelines, existing large-scale methods lack semantic awareness and support for parallel instance-level training, and Gaussian-based lightweight building modeling remains largely unexplored. In this paper, we propose SemLite-GS, a building-centric Gaussian splatting framework, which combines semantic selection, parallel training, and plane-constrained mesh reconstruction to generate lightweight building models from large-scale UAV scenarios. First, a semantic orthographic-view reference map  is generated to establish globally consistent building identities. These identities are propagated to the input oblique images to extract camera--instance pairs and corresponding regions of interest (ROIs) for instance-level optimization. Second, a semantic-selective parallel PGSR rasterizer processes camera--instance pairs from multiple buildings and views concurrently while restricting computation to relevant image tiles and associated Gaussian primitives. Surrounding contextual Gaussians are retained for visibility consistency but excluded from parameter optimization, thereby saving unnecessary computation. Third, we propose a plane-guided SDF (Signed Distance Function) refinement strategy for lightweight mesh extraction. Architectural planes, extracted from rendered depth, opacity, and normal maps, guide SDF optimization to directly generate lightweight meshes that preserve sharp architectural boundaries. Furthermore, polygon density is adaptively allocated according to geometric complexity, minimizing  redundant tessellation in planar regions while preserving sharp architectural transitions. Experiments on multiple large-scale UAV datasets demonstrate that our SemLite-GS substantially reduces building reconstruction time compared with representative full-scene 3DGS methods. With comparable polygon budgets, SemLite-GS achieves superior preservation of dominant planar structures and sharp architectural transitions compared with conventional dense reconstruction followed by mesh decimation. The source code is available at \textit{https://github.com/IRIS-LAB-whu/SemLite-GS}.
\end{abstract}


\maketitle

\section{Introduction}
\label{sec:introduction}

Lightweight 3D building models, represented as compact, low-poly meshes typically comprising several hundred faces \citep{liu2025gs2poly}, are essential for efficient urban-scale representation and spatial analysis \citep{peters2022automated}. Serving as foundational 3D primitives, they support a wide range of applications, such as digital twins, geographic information systems, and large-scale visualization \citep{biljecki2015applications,schuegraf2025sat2building}. Compared to dense point clouds and highly tessellated meshes, lightweight 3D building models mainly capture dominant building structures using far fewer vertices and faces, thereby significantly reducing storage, transmission, and rendering costs. Over the past few decades, UAV oblique photogrammetry has been widely studied and has enabled detailed 3D reconstruction of large-scale urban environments. Despite these advances, many practical applications require only a specific subset of target buildings rather than complete scene models. Consequently, efficiently reconstructing these target buildings from large UAV image blocks remains a critical challenge.

\begin{figure*}[!t]
\centering
\includegraphics[width=\textwidth]{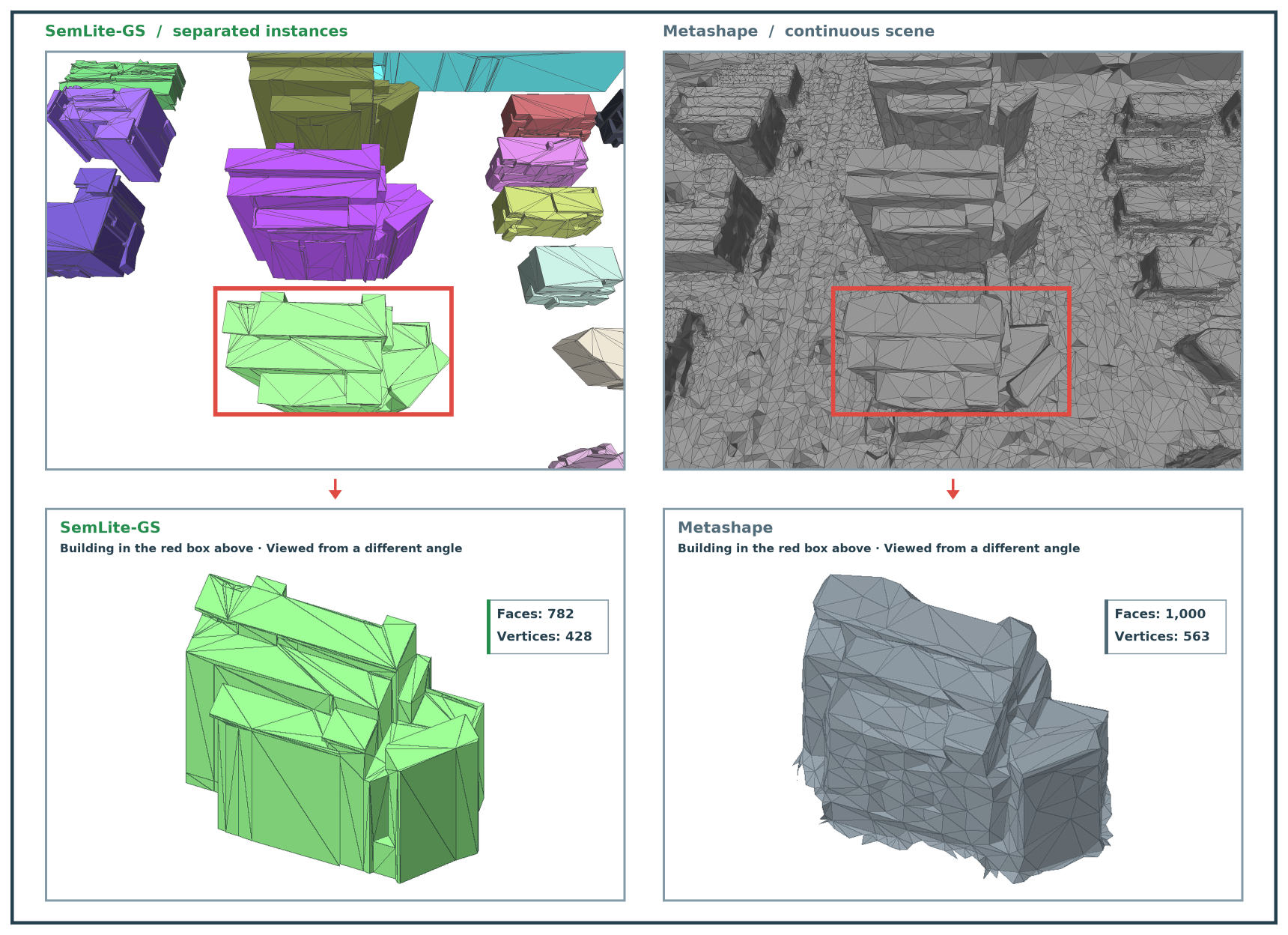}
\caption{Qualitative comparison of building meshes generated by SemLite-GS and Metashape \citep{metashape}. The upper panels show individually identified building meshes from SemLite-GS and a continuous scene mesh from Metashape. The lower panels show the building marked by the red boxes from a different oblique viewpoint, highlighting differences in planar triangulation and building-edge regularity.}
\label{fig:workflow_comparison}
\end{figure*}

Conventional photogrammetric workflows typically address lightweight building modeling through a sequential pipeline: dense scene reconstruction, followed by building extraction, and finally mesh simplification ~\citep{garland1997surface,bouzas2020structure,li2024shapepreserving,bauchet2020kinetic,sulzer2024compod}. When applied to target-building reconstruction from large UAV image blocks, however, this paradigm exhibits two limitations. First, dense reconstruction is performed over the entire scene, including vast non-target regions, resulting in unnecessary computational overhead. Second, lightweight modeling is treated merely as a post-processing operation after dense mesh generation. Although mesh simplification reduces polygon count, aggressive decimation often compromises architectural structures, such as roof ridges, corners, and facade intersections. These limitations motivate the development of a novel reconstruction paradigm that integrates semantic target selection into optimization and explicitly incorporates architectural structures into lightweight model generation. Figure~\ref{fig:workflow_comparison} compares building meshes generated by the commercial photogrammetric software Metashape \citep{metashape} and the proposed SemLite-GS framework, illustrating that SemLite-GS represents planar surfaces with fewer triangles, yields straighter and more regular building edges, and reconstructs individually identified building instances.

Recent advances in 3D Gaussian Splatting (3DGS) provide a promising alternative to revisit conventional photogrammetric workflows. The explicit nature of Gaussian primitives and differentiable rasterization enables the selective optimization of scene components by controlling the participating primitives \citep{kerbl2023gaussians}. Furthermore, surface reconstruction extensions, such as 2D Gaussian Splatting \citep{huang2024twodgs} and PGSR \citep{chen2025pgsr}, provide rich geometric cues for extracting architectural structures from rendered depth and normal information \citep{dai2024gaussiansurfels,huang2024twodgs}. These properties highlight the potential of 3DGS for building-specific optimization and geometry-guided lightweight mesh generation. However, existing large-scale 3D Gaussian Splatting methods mainly focus on scene-level reconstruction based on  spatial partitioning rather than semantic building identities \citep{kerbl2024hierarchical,lin2024vastgaussian,liu2024citygaussian,gao2025citygsx,liu2025citygaussianv2}, thus failing to directly support identity-conditioned parallel training of multiple buildings. Moreover, recent advances in Gaussian-based surface reconstruction \citep{chen2025pgsr,yu2024gof,lyu2024threedgsr, wang2025pgsag} mainly focus on improving surface recovery quality but do not address lightweight model generation under constrained polygon budgets. Accurate surface reconstruction alone does not dictate how polygons should be distributed to preserve critical architectural structures.

Several recent studies have explored lightweight architectural modeling based on Gaussian representations. For instance, GS2Poly \citep{liu2025gs2poly} integrates Gaussian-derived planar primitives with opacity field-guided spatial partitioning to generate compact polygonal building models, while SF-Recon \citep{li2025sfrecon} exploits informative Gaussians for edge pruning and Delaunay triangulation to construct lightweight building meshes. Although these studies demonstrate the potential of Gaussian representations for compact architectural modeling, they generally assume that the target building regions or corresponding Gaussian representations have already been isolated prior to mesh reconstruction. Consequently, they mainly focus on building-level lightweight model generation from pre-extracted inputs, rather than addressing the complete process of automatically identifying target buildings from large UAV image blocks, efficiently optimizing their corresponding Gaussian representations, and generating structure-preserving lightweight meshes.

In this paper, we propose SemLite-GS, a semantic-selective parallel Gaussian Splatting framework featuring structure-aware lightweight mesh reconstruction for large UAV oblique image blocks. Starting from registered multi-view UAV images with known camera parameters, SemLite-GS rapidly initializes a shared scene-level Gaussian representation, constructs an orthographic building-instance map, and formulates each building identity as an explicit reconstruction target. The orthographic semantic information is propagated to the oblique views to establish camera–instance correspondences and define regions of interest (ROIs) for optimization. Based on these correspondences, a semantic-selective parallel PGSR optimization strategy is introduced, wherein multiple camera--instance pairs are processed in parallel to refine the building instances through ROI-constrained rasterization and target-selective parameter updates. While target Gaussians receive direct optimization signals, contextual Gaussians are retained solely for visibility consistency. This mechanism enables efficient parallel reconstruction without the need to repeatedly optimize the entire scene. Furthermore, architectural planes extracted from the optimized Gaussian representation are incorporated as geometric priors into SDF refinement, facilitating the direct generation of lightweight meshes with adaptive polygon allocation and preserved structural transitions. By integrating semantic selection during Gaussian optimization and structural guidance during mesh reconstruction, SemLite-GS effectively bypasses dense scene reconstruction followed by generic mesh decimation.

The main contributions of this work are summarized as follows:

\begin{itemize}

\item \textbf{Semantic-selective Gaussian reconstruction for target-oriented building modeling}: A semantic-selective Gaussian Splatting framework is proposed to formulate individual buildings as explicit reconstruction targets. By propagating orthographic semantic identities to oblique views, the framework establishes a camera--instance observation graph and enables building-specific Gaussian optimization.

\item \textbf{ROI-aware parallel PGSR optimization with contextual visibility preservation}: An efficient instance-level Gaussian training strategy is introduced to process multiple camera--instance pairs in parallel within a shared GPU workload. ROI-local rasterization and target-conditioned Gaussian updates reduce unnecessary computation while retaining fixed contextual Gaussians for visibility reasoning.

\item \textbf{Plane-guided Gaussian-to-mesh reconstruction for lightweight structural modeling}: A plane-guided mesh generation strategy is developed by incorporating architectural planes into SDF refinement. The proposed strategy enables adaptive polygon allocation and preserves essential architectural structures during lightweight model generation.
    
\end{itemize}

\section{Related Work}
\label{sec:related_work}

\subsection{Lightweight Building Modeling}
\label{sec:rw_lightweight}

Lightweight building modeling aims to generate compact geometric representations while preserving dominant architectural structures under constrained polygon budgets. Existing approaches can be broadly categorized into three main streams: dense-mesh simplification, primitive-based reconstruction, and learning-based polygon generation.

Dense-mesh simplification, a widely adopted strategy, typically reconstructs detailed geometry through structure-from-motion, multi-view stereo, point-cloud fusion, and surface meshing, followed by polygon reduction \citep{schonberger2016sfm,schonberger2016mvs}. Classical methods, such as quadric-error metric optimization \citep{garland1997surface} and variational geometric proxies \citep{cohensteiner2004variational}, simplify meshes by minimizing geometric approximation errors. Recent studies further incorporate structural constraints, including planar adjacency preservation \citep{bouzas2020structure} and structure-aware edge collapse \citep{li2024shapepreserving}, to better retain architectural features. However, these approaches inherently rely on the availability and quality of dense geometric reconstruction; consequently, geometric errors introduced prior to simplification cannot be recovered during the polygon reduction phase.

Primitive-based methods directly construct compact building models from geometric abstractions, such as planes, lines, contours, or spatial cells. Manhattan-world models represent regular buildings using axis-aligned primitives \citep{li2016manhattan}, while PolyFit and TopoLAP exploit planar and linear relationships to generate watertight polygonal models \citep{nan2017polyfit,xie2019topolap}. More recent approaches extend primitive-based reconstruction to complex urban scenes through adaptive plane arrangements, hypothesis selection, and spatial partitioning \citep{bauchet2020kinetic,fang2020connect,huang2022city3d,chen2022compact,sulzer2024compod,bauchet2024simplicity}. Although these methods can generate compact and structurally meaningful models, they generally require reliable geometric observations and explicit primitive extraction as prerequisites for model generation.

Learning-based methods infer compact building structures directly from geometric observations. For instance, Point2Roof predicts roof structures from airborne point clouds \citep{woldesenbet2022point2roof}, while Point2Building autoregressively generates polygonal vertices and faces \citep{liu2024point2building}. While learning-based approaches reduce dependence on manually designed geometric rules, they usually require instance-level geometric inputs and may suffer from domain gaps when generalizing from training data to real-world UAV scenes.

Overall, existing lightweight building reconstruction paradigms predominantly  focus on simplifying or generating compact geometry after suitable building observations have been obtained. They typically assume that target buildings have already been isolated or that reliable point clouds and meshes are available. Therefore, they do not directly address selective reconstruction of target buildings from large UAV image blocks prior to the lightweight mesh generation phase.

\subsection{Gaussian-Based Surface Reconstruction}
\label{sec:rw_gs_geometry}

Recent advancements in Gaussian-based surface reconstruction have primarily  focused on improving geometric fidelity and recovering accurate surfaces from Gaussian representations. Existing studies enhance Gaussian-based geometry by introducing surface-aware representations, geometric regularization, and additional geometric priors, or by extracting explicit surfaces from optimized Gaussian representations. These advances have substantially improved the capability of Gaussian models, enabling high-quality surface modeling beyond the original view-synthesis-oriented 3D Gaussian Splatting.

Representative methods enhance Gaussian geometry from diverse perspectives. SuGaR encourages Gaussians to align with underlying scene surfaces and subsequently applies Poisson reconstruction for mesh extraction \citep{guedon2024sugar}. Gaussian Surfels introduces flattened elliptical primitives for local surface representation \citep{dai2024gaussiansurfels}, while PGSR flattens anisotropic 3D Gaussians toward local planes, renders plane-distance and normal maps for depth recovery, and uses single-view geometric and multi-view photometric and geometric constraints to improve surface reconstruction across varying viewpoints \citep{chen2025pgsr}. Further studies incorporate additional geometric priors, including depth and normal supervision \citep{turkulainen2025dnsplatter,zhang2024radegs}, to enhance geometric consistency and surface recovery quality. Beyond refining Gaussian geometry, several methods have investigated explicit surface or mesh reconstruction from Gaussian representations. Gaussian Opacity Fields extracts surfaces through level-set extraction from continuous opacity fields \citep{yu2024gof}, while 3DGSR jointly optimizes Gaussian representations and signed distance fields for surface reconstruction \citep{lyu2024threedgsr}. Geometry Field Splatting reconstructs surfaces from Gaussian-induced geometry fields \citep{jiang2025geometry}, and GS2Mesh generates explicit meshes by rendering and fusing stereo depth observations \citep{wolf2024gs2mesh}. These methods significantly improve Gaussian-based geometric reconstruction and demonstrate the potential of Gaussian representations for explicit surface modeling. However, their primary objective remains high-fidelity surface recovery, focusing on preserving detailed geometry rather than generating lightweight models with structure-aware simplification and explicit polygon budget control. Nevertheless, their success has further motivated the application of Gaussian representations to structured environments, particularly architectural reconstruction scenarios.

Building upon these foundational advances, recent studies have extended Gaussian-based reconstruction specifically to architectural modeling. Gaussian Building Mesh combines semantic masks with Gaussian geometry for building mesh generation \citep{gao2024gbm}, while GS4Buildings introduces LoD2-derived depth and normal priors to improve building reconstruction \citep{zhang2025gs4buildings}. GS2Poly integrates Gaussian-derived planar primitives with opacity-field-guided spatial partitioning \citep{liu2025gs2poly}, and SF-Recon selects informative Gaussians for edge pruning and Delaunay triangulation \citep{li2025sfrecon}. These methods demonstrate the feasibility of Gaussian-based architectural reconstruction. However, they generally assume that Gaussian representations for target regions are already available and primarily focus on geometric recovery or mesh extraction. The problem of jointly selecting target buildings from large UAV scenes and generating lightweight meshes with structure-aware polygon allocation remains insufficiently explored.

\subsection{Scalable Gaussian Splatting}
\label{sec:rw_scalable}

Scalable Gaussian Splatting methods aim to mitigate primitive redundancy, memory consumption, and optimization costs when extending 3DGS to large-scale scenes. Existing approaches mainly improve scalability through structured or compressed representations and scene decomposition strategies.

Structured representation methods organize Gaussians according to spatial locations or levels of detail. Scaffold-GS generates view-adaptive Gaussians from anchor points \citep{lu2024scaffoldgs}, while Octree-GS introduces an octree-based level-of-detail hierarchy for multiscale representation \citep{ren2024octreegs}. Hierarchical 3D Gaussians combine independently optimized scene chunks into a multiresolution representation \citep{kerbl2024hierarchical}. Other approaches directly reduce representation redundancy. LightGaussian combines pruning, distillation, and quantization to compress Gaussian representations \citep{fan2024lightgaussian}; CompGS encodes Gaussian parameters using compact codebooks \citep{navaneet2024compgs}; and EfficientGS controls primitive growth through selective densification and pruning \citep{liu2025efficientgs}. Although these methods effectively reduce representation or rendering costs, they generally retain the complete scene as the reconstruction domain.

Scene-decomposition methods enhance scalability by dividing large scenes into manageable reconstruction units. VastGaussian partitions scenes into visibility-aware cells \citep{lin2024vastgaussian}, while CityGaussian combines block-wise training with level-of-detail representations for large-scale urban reconstruction \citep{liu2024citygaussian}. CityGaussianV2 and CityGS-X further improve parallel optimization and geometric supervision for city-scale scenes \citep{liu2025citygaussianv2,gao2025citygsx}. However, this parallelism is organized according to spatial coverage rather than semantic building identities, such that a single reconstruction unit may contain both target buildings and irrelevant surroundings.

To introduce semantic awareness, recent approaches have incorporated building masks to guide Gaussian optimization \citep{wang2025pgsag, Proxy-coupled}. For example, PG-SAG refines view-specific masks via multi-view voting and performs masked optimization of visible building regions \citep{wang2025pgsag}. Although recent feature-embedded Gaussian methods have also explored semantic associations \citep{qin2024langsplat, cen2025saga, song2026pointgs, liu2026gensplat}, they primarily focus on downstream segmentation rather than reconstruction optimization. Furthermore, establishing globally consistent instance associations across oblique UAV views and enabling the efficient parallel optimization of multiple building targets remain significant challenges. To bridge this gap, SemLite-GS leverages an orthographic instance map as a globally consistent semantic anchor, propagating building identities to oblique views to establish camera-instance pairs. In contrast to previous methods that optimize entire spatial blocks indiscriminately, our approach enables targeted, instance-level parallel optimization while strictly preserving contextual visibility consistency.

\section{Preliminaries}
\label{sec:preliminaries}

3D Gaussian Splatting (3DGS) represents a scene as a collection of explicit anisotropic Gaussian primitives \citep{kerbl2023gaussians}. Each primitive $g_i$ is parameterized by a center $\boldsymbol{\mu}_i$, a covariance matrix $\boldsymbol{\Sigma}_i$, a base opacity $\alpha_i$, and spherical harmonic coefficients describing view-dependent appearance. The covariance determines the spatial orientation and anisotropic extent of the primitive and is parameterized to ensure positive semidefiniteness.

During rendering, the Gaussian primitives are projected onto the image plane using a local affine approximation of perspective projection. For pixel $\mathbf{p}$ in camera $c$, let $\mathcal{I}_c(\mathbf{p})$ denote the depth-ordered indices of the Gaussians whose projected footprints cover $\mathbf{p}$. The transmittance preceding Gaussian $g_i$ and the rendered color are given by

\begin{equation}
\begin{aligned}
T_{ci}(\mathbf{p})
&=
\prod_{\substack{j\in\mathcal{I}_c(\mathbf{p})\\j\prec i}}
\left[
1-\alpha_{cj}(\mathbf{p})
\right],\\
\widehat{\mathbf{C}}_c(\mathbf{p})
&=
\sum_{i\in\mathcal{I}_c(\mathbf{p})}
T_{ci}(\mathbf{p})\,
\alpha_{ci}(\mathbf{p})\,
\mathbf{c}_{ci},
\end{aligned}
\label{eq:prelim_compositing}
\end{equation}

where $j\prec i$ indicates that $g_j$ precedes $g_i$ along the viewing direction, $\mathbf{c}_{ci}$ denotes the view-dependent color of $g_i$, and $\alpha_{ci}(\mathbf{p})$ is its effective opacity at $\mathbf{p}$, obtained by combining the learnable base opacity $\alpha_i$ with the projected Gaussian response. The background contribution is omitted for clarity.

The rasterizer organizes this computation into screen-space tiles. Each projected Gaussian is duplicated for the tiles intersected by its footprint, after which the resulting Gaussian--tile entries are sorted by tile index and depth. Each nonempty depth-ordered tile range is then assigned to a CUDA thread block for pixel-wise compositing. This organization provides the computational basis for the ROI-constrained batched rasterizer described in Sec.~\ref{sec:batched_pgsr}.

Although 3DGS is effective for novel-view synthesis, its ability to recover accurate surface geometry remains limited. Planar-based Gaussian Splatting Reconstruction (PGSR) is a Gaussian-based method developed for surface reconstruction that encourages anisotropic 3D Gaussians to approximate local planes \citep{chen2025pgsr}. For Gaussian $g_i$ with rotation $\mathbf{R}_i$ and positive scales $s_{i,1},s_{i,2},s_{i,3}$, its covariance is
\begin{equation}
\boldsymbol{\Sigma}_i=\mathbf{R}_i
\operatorname{diag}(s_{i,1}^{2},s_{i,2}^{2},s_{i,3}^{2})
\mathbf{R}_i^{\mathsf T}
\label{eq:prelim_pgsr_covariance}
\end{equation}
The shortest scale direction defines the normal of the associated local plane, while minimum-scale regularization encourages the Gaussian to flatten toward that plane.

For camera $c$, let $\mathbf{n}_{ci}$ be the unit normal of the
Gaussian-associated plane in camera coordinates, and let
$d_{ci}=\mathbf{n}_{ci}^{\mathsf T}\boldsymbol{\mu}_{ci}$ be its signed
distance from the camera origin, where $\boldsymbol{\mu}_{ci}$ is the
Gaussian center in camera coordinates. With compositing weight
$w_{ci}(\mathbf{p})=T_{ci}(\mathbf{p})\alpha_{ci}(\mathbf{p})$,
PGSR renders the blended normal $\widehat{\mathbf{N}}_c$ and plane
distance $\widehat{\mathcal{D}}_c$:

\begin{equation}
\widehat{\mathbf{N}}_c(\mathbf{p})
=\sum_{i\in\mathcal{I}_c(\mathbf{p})}w_{ci}(\mathbf{p})\mathbf{n}_{ci},
\qquad
\widehat{\mathcal{D}}_c(\mathbf{p})
=\sum_{i\in\mathcal{I}_c(\mathbf{p})}w_{ci}(\mathbf{p})d_{ci}
\label{eq:prelim_pgsr_plane_maps}
\end{equation}
The plane distance $\widehat{\mathcal{D}}_c(\mathbf{p})$ is distinct
from the plane-derived depth $\widehat D_c(\mathbf{p})$. For intrinsic
matrix $\mathbf{K}_c$ and homogeneous pixel coordinate
$\widetilde{\mathbf{p}}$, the latter is given by

\begin{equation}
\widehat{D}_c(\mathbf{p})
=\frac{\widehat{\mathcal{D}}_c(\mathbf{p})}
{\widehat{\mathbf{N}}_c(\mathbf{p})^{\mathsf T}
\mathbf{K}_c^{-1}\widetilde{\mathbf{p}}}
\label{eq:prelim_pgsr_depth}
\end{equation}

PGSR groups single-view geometric, multi-view photometric, and multi-view geometric consistency terms into the following regularization loss:
\begin{equation}
\mathcal{L}_{\mathrm{geom}}
=\lambda_{sv}\mathcal{L}_{\mathrm{svgeom}}
+\lambda_{mp}\mathcal{L}_{\mathrm{mvphoto}}
+\lambda_{mg}\mathcal{L}_{\mathrm{mvgeom}}.
\label{eq:prelim_pgsr_geom}
\end{equation}
The overall objective is
\begin{equation}
\mathcal{L}_{\mathrm{PGSR}}
=\mathcal{L}_{\mathrm{rgb}}
+\lambda_1\mathcal{L}_{s}
+\mathcal{L}_{\mathrm{geom}},
\label{eq:prelim_objective}
\end{equation}
where $\mathcal{L}_{\mathrm{rgb}}
=(1-\lambda_s)\mathcal{L}_1
+\lambda_s\mathcal{L}_{\mathrm{D\text{-}SSIM}}$ combines pixel-wise and structural photometric supervision. The minimum-scale loss $\mathcal{L}_{s}$ encourages Gaussians to flatten toward local planes. The single-view term $\mathcal{L}_{\mathrm{svgeom}}$ encourages agreement between rendered normals and normals derived from rendered depth. The multi-view photometric term $\mathcal{L}_{\mathrm{mvphoto}}$ measures the consistency of image patches matched through the reconstructed planes, while $\mathcal{L}_{\mathrm{mvgeom}}$ penalizes geometric inconsistency under forward and backward reprojection. The coefficients $\lambda_s$, $\lambda_1$, $\lambda_{sv}$, $\lambda_{mp}$, and $\lambda_{mg}$ balance the corresponding terms.

\section{Methods}
\label{sec:methods}

\subsection{Overview}
\label{sec:method_overview}

Given registered UAV oblique images and camera parameters, SemLite-GS reconstructs lightweight meshes for selected building instances within a large scene. A rapidly initialized scene-level Gaussian representation provides a shared reference for semantic association and visibility reasoning, while subsequent training and mesh reconstruction focus on the identified target buildings. Building identity thus defines the optimization domain, allowing the Gaussian representations of multiple buildings to be trained in parallel within a shared GPU workload without updating unrelated scene regions.

\begin{figure*}[!tp]
\centering
\includegraphics[width=\linewidth]{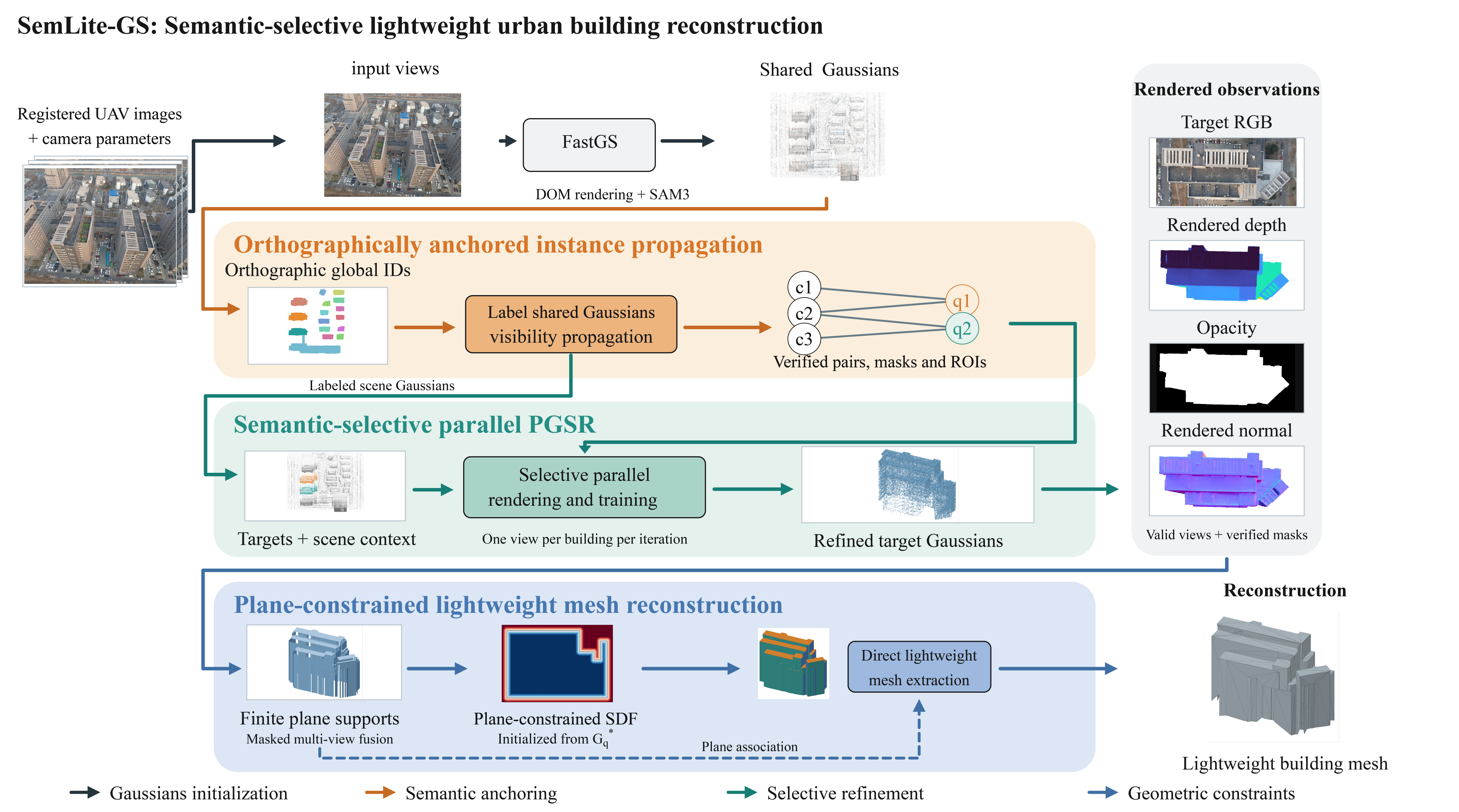}
\caption{Overview of SemLite-GS. A FastGS initializes a shared scene-level Gaussian representation from registered UAV images and their camera parameters. Orthographically anchored instance propagation establishes globally consistent building identities and constructs a camera--instance observation graph with verified masks and ROIs. Semantic-selective parallel PGSR refines multiple building instances concurrently within target-specific Gaussian and image domains, retaining contextual Gaussians for visibility-consistent rendering while updating only the corresponding targets. The refined Gaussians provide multi-view RGB, depth, opacity, and normal observations. Finite architectural planes estimated from the masked geometric observations constrain Gaussian-initialized SDFs and guide plane-aware surface consolidation and triangulation to produce lightweight building meshes. Arrow colors distinguish scene and data flow, semantic anchoring, selective training, and geometric constraints; the dashed arrow indicates the reuse of finite plane supports during mesh extraction.}
\label{fig:method_overview}
\end{figure*}

As illustrated in Fig.~\ref{fig:method_overview}, SemLite-GS comprises three stages. First, based on FastGS and SAM 3, Sec.~\ref{sec:semantic_propagation} establishes building identities in a global orthographic reference, assigns them to the Gaussians of the whole scene, and propagates their visible support to the original oblique views. This stage produces a camera--instance observation graph containing the verified masks and regions of interest (ROIs) of the building instance. Second, Sec.~\ref{sec:batched_pgsr} introduces a semantic-selective parallel PGSR rasterizer that concurrently processes camera--instance pairs from multiple buildings and views within ROI-constrained rendering domains. Target Gaussians of buildings are optimized, while fixed contextual Gaussians are retained for visibility-aware rendering. Finally, Sec.~\ref{sec:mesh_generation} extracts finite architectural planes from the optimized Gaussian representations, which are employed to constrain instance-level signed distance fields. The optimized fields are converted into lightweight meshes by simplifying planar interiors while preserving ridges, corners, and plane intersections.

\subsection{Orthographically Anchored Instance Propagation}
\label{sec:semantic_propagation}

\begin{figure*}[!tp]
\centering
\includegraphics[width=\linewidth]{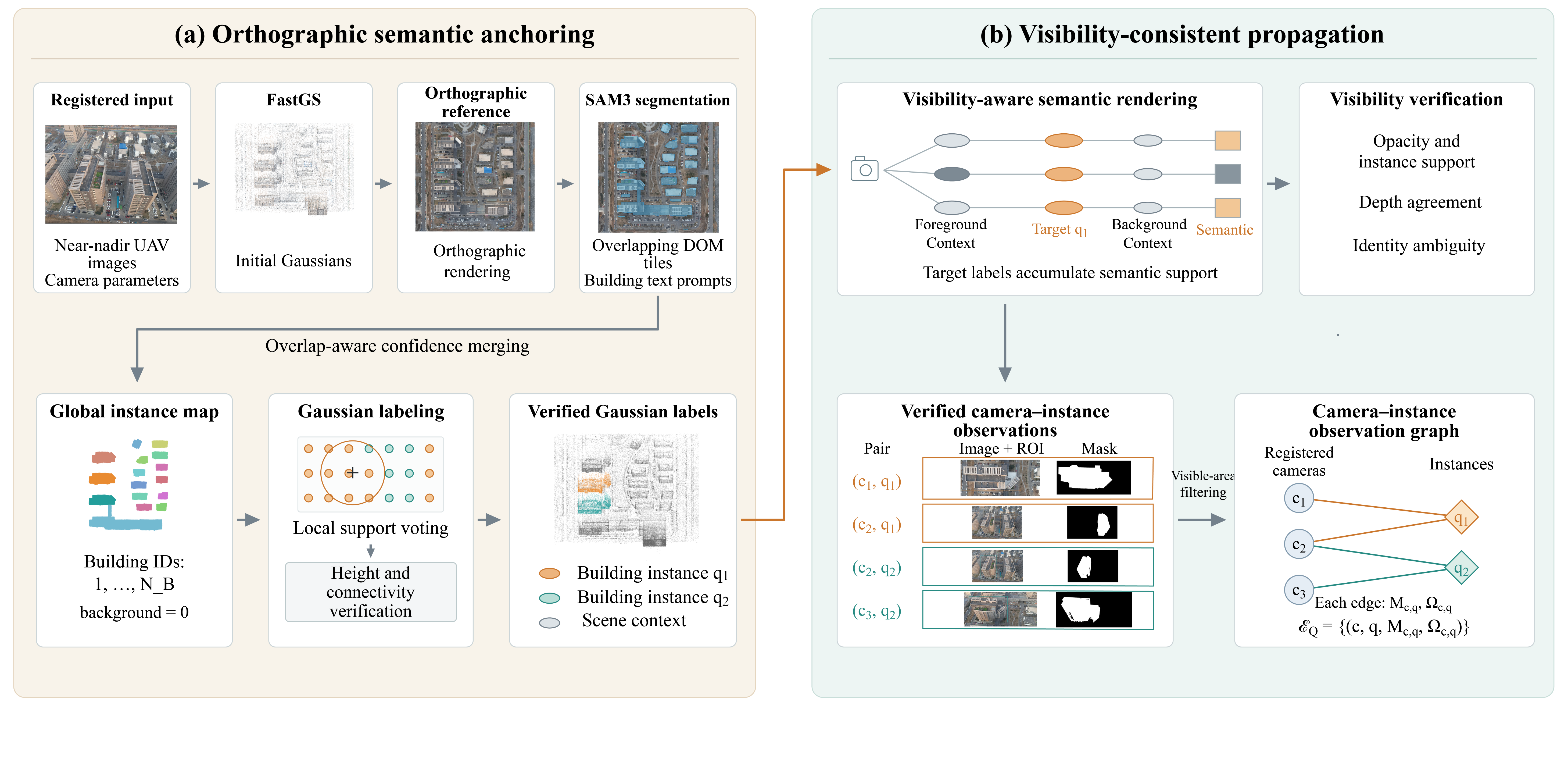}
\caption{Workflow of orthographically anchored instance propagation. (a) Orthographic semantic anchoring: the FastGS initializes a shared Gaussian representation for DOM rendering. SAM3 predictions from overlapping DOM tiles are merged into a global instance map, and building identities are assigned to Gaussians through local support voting and verification of height and connectivity. (b) Visibility-consistent propagation: semantic rendering accounts for full-scene occlusion, with all Gaussians determining transmittance and target-labeled Gaussians contributing instance support. Visibility verification and mask refinement produce camera--instance observations, which are retained through visible-area filtering to construct the observation graph. Each edge stores a verified mask $M_{c,q}$ and its associated ROI $\Omega_{c,q}$.}
\label{fig:instance_propagation_pipeline}
\end{figure*}

Applying SAM3 \citep{carion2026sam} independently to individual views can yield inconsistent mask boundaries and instance assignments for the same building because of illumination changes, viewpoint variations, and prediction uncertainty. Large buildings may also be only partially observed or severely occluded in various individual images, resulting in incomplete or fragmented masks. As illustrated in Fig.~\ref{fig:instance_propagation_pipeline}, to obtain consistent cross-view supervision, SemLite-GS first establishes a global building-instance reference in the orthographic domain and assigns the resulting identities to the shared scene-level Gaussian representation. The identity-labeled Gaussians are subsequently rendered into the original oblique views using visibility-aware compositing of the complete scene. This procedure maintains consistent building identities across views, accounts for occlusion by surrounding scene elements, and generates verified masks, ROIs, and camera--instance pairs required for subsequent selective optimization.

\subsubsection{Orthographic Semantic Anchoring}
\label{sec:orthographic_anchoring}

FastGS \citep{ren2026fastgs} is trained on the registered UAV images to efficiently initialize the scene-level Gaussian representation $\mathcal{G}^{0}$. The imagery provides broad and relatively uniform coverage for subsequent DOM generation. For subsequent PGSR training, the initialized representation retains each Gaussian's center, rotation, three scale parameters, opacity, and appearance attributes.

Assuming that the world $z$-axis represents the vertical direction, a downward-looking virtual camera is defined in the scene coordinate system \citep{wang2026tortho}. Following the orthographic rendering formulation of Tortho-Gaussian \citep{wang2026tortho}, $\mathcal{G}^{0}$ is rendered from this virtual camera to produce the digital orthophoto map (DOM). The camera position and spatial extent are estimated from robust bounds of the Gaussian centers $\boldsymbol{\mu}_i$, reducing sensitivity to distant or floating outliers. The output resolution is specified according to the desired ground sampling distance. For large scenes, rendering is performed over overlapping tiles to control peak memory consumption while maintaining complete coverage across tile boundaries.

Let $\pi_o(\boldsymbol{\mu}_i)$ denote the projection of Gaussian center $\boldsymbol{\mu}_i$ onto the DOM. This projection establishes a fixed correspondence between the DOM and the scene coordinate system. Depth is retained for visibility ordering but does not affect the spatial scale of the orthographic image. Gaussian appearance is evaluated using low-order spherical harmonics to limit excessive view-dependent variation under the virtual camera.

Building instances are extracted from each DOM tile using SAM3 \citep{carion2026sam} conditioned on building-related text prompts. The predicted masks are transformed into the common DOM coordinate system and merged according to spatial overlap and prediction confidence. Low-confidence predictions and small disconnected regions are removed, and locally supported holes are filled to improve mask completeness. Each retained building is assigned a unique identity $q\in\{1,\ldots,N_B\}$, producing the orthographic instance map $L^o$, where $L^o(\mathbf{p})=0$ denotes background. The subset of identities requested by the reconstruction task is denoted by $\mathcal{Q}$.

The identities in $L^o$ are subsequently assigned to the scene Gaussians. To reduce sensitivity to projection errors and uncertain mask boundaries, the assignment is determined from the labels within a circular neighborhood $\mathcal{N}_i$ centered at $\pi_o(\boldsymbol{\mu}_i)$. Let $v_{i,q}=\sum_{\mathbf{p}\in\mathcal{N}_i}\mathbb{I}[L^o(\mathbf{p})=q]$ denote the support received by building $q$. The instance label of Gaussian $g_i$ is assigned as

\begin{equation}
\ell_i
=
\begin{cases}
\displaystyle
\arg\max_{q\in\{1,\ldots,N_B\}} v_{i,q},
&
\displaystyle
\max_q v_{i,q}\geq\tau_n
\ \text{and}\
\frac{\max_q v_{i,q}}{|\mathcal{N}_i|}
\geq\tau_r,\\[9pt]
0,
&
\text{otherwise},
\end{cases}
\label{eq:gaussian_instance_assignment}
\end{equation}

Here, $\mathcal{N}_i$ contains five DOM pixels sampled
within a one-pixel radius. The most frequent non-background label is
assigned whenever at least one sampled pixel belongs to a building.
For the threshold formulation in Eq.~\ref{eq:gaussian_instance_assignment},
this rule corresponds to $\tau_n=1$ and $\tau_r=0.2$; the two thresholds
are not independently tuned. An assignment of $\ell_i=0$ denotes an
unlabeled Gaussian.

As the orthographic assignment primarily constrains horizontal location, the initial labels are further verified in three dimensions. Robust height bounds remove vertically inconsistent Gaussians, while voxel-based connectivity retains the principal component of each building and suppresses isolated vegetation, terrain, and floating structures. Rejected Gaussians remain in $\mathcal{G}^{0}$ as scene context but provide no semantic support for any building instance.

\subsubsection{Visibility-Consistent Propagation}
\label{sec:visibility_consistent_rendering}

The obtained Gaussian identities are propagated to the original oblique images through semantic rendering. The semantic rasterizer follows the same projection, depth ordering, and front-to-back compositing as used in appearance rendering, allowing the propagated labels to respect full-scene visibility rather than the projection of an isolated building model.

For building $q$, its visibility-weighted support at pixel $\mathbf{p}$ in camera $c$ is defined as

\begin{equation}
S_{c,q}(\mathbf{p})
=
\sum_{i\in\mathcal{I}_c(\mathbf{p})}
\mathbb{I}[\ell_i=q]\,
\mathbb{I}[\alpha_{ci}(\mathbf{p})\geq\tau_{\mathrm{sem}}]\,
T_{ci}(\mathbf{p})\,
\alpha_{ci}(\mathbf{p}),
\label{eq:visibility_weighted_instance_support}
\end{equation}

where $\tau_{\mathrm{sem}}$ suppresses weak contributions from Gaussian tails and $T_{ci}(\mathbf{p})$ is the full-scene transmittance defined in Eq.~\ref{eq:prelim_compositing}. Only Gaussians labeled as building $q$ contribute semantic support, whereas all preceding Gaussians, including terrain, vegetation, and neighboring buildings, contribute to transmittance. Foreground objects can therefore occlude a building without being incorporated into its semantic mask.

For each pixel, the building with the highest visibility-weighted support is selected as the initial candidate. Candidates with insufficient scene opacity, weak instance support, or ambiguous contributions from multiple buildings are rejected. A boundary-aware criterion based on the strongest visible single-Gaussian contribution retains valid pixels along thin roof and facade structures. Finally, target-building opacity and depth are evaluated against the full-scene visible depth to discard candidate pixels where the building is occluded by a foreground surface.

The verified masks are further refined by removing small disconnected components and filling locally supported holes. The resulting binary mask of building $q$ in camera $c$ is denoted by $M_{c,q}$. Its foreground bounding box is expanded by a small margin to define the ROI $\Omega_{c,q}$, retaining the contextual information required for visibility reasoning near building boundaries. The generated masks and ROIs are subsequently aligned with the image resolution and padding configuration used during training. Observations with sufficient visible support define the edge set of the camera--instance graph:

\begin{equation}
\mathcal{E}_{\mathcal{Q}}
=
\left\{
(c,q,M_{c,q},\Omega_{c,q})
\;\middle|\;
q\in\mathcal{Q},\;
\operatorname{area}(M_{c,q})\geq\tau_{\mathrm{area}}
\right\},
\label{eq:camera_instance_graph}
\end{equation}

where $\tau_{\mathrm{area}}$ denotes the minimum visible-area threshold. Views in which a target building is too distant, severely occluded, or represented by too few pixels are consequently excluded.

Each edge in $\mathcal{E}_{\mathcal{Q}}$ associates a globally consistent building identity with one visibility-verified camera observation and stores its view-dependent mask and ROI. The propagation stage thus converts the orthographic identity reference into a sparse, task-conditioned set of multi-view observations. These graph edges are used directly as the supervision and computation units of the semantic-selective optimization described in Sec.~\ref{sec:batched_pgsr}.

\subsection{Semantic-Selective Parallel PGSR}
\label{sec:batched_pgsr}

\begin{figure*}[!tp]
\centering
\includegraphics[width=\linewidth]{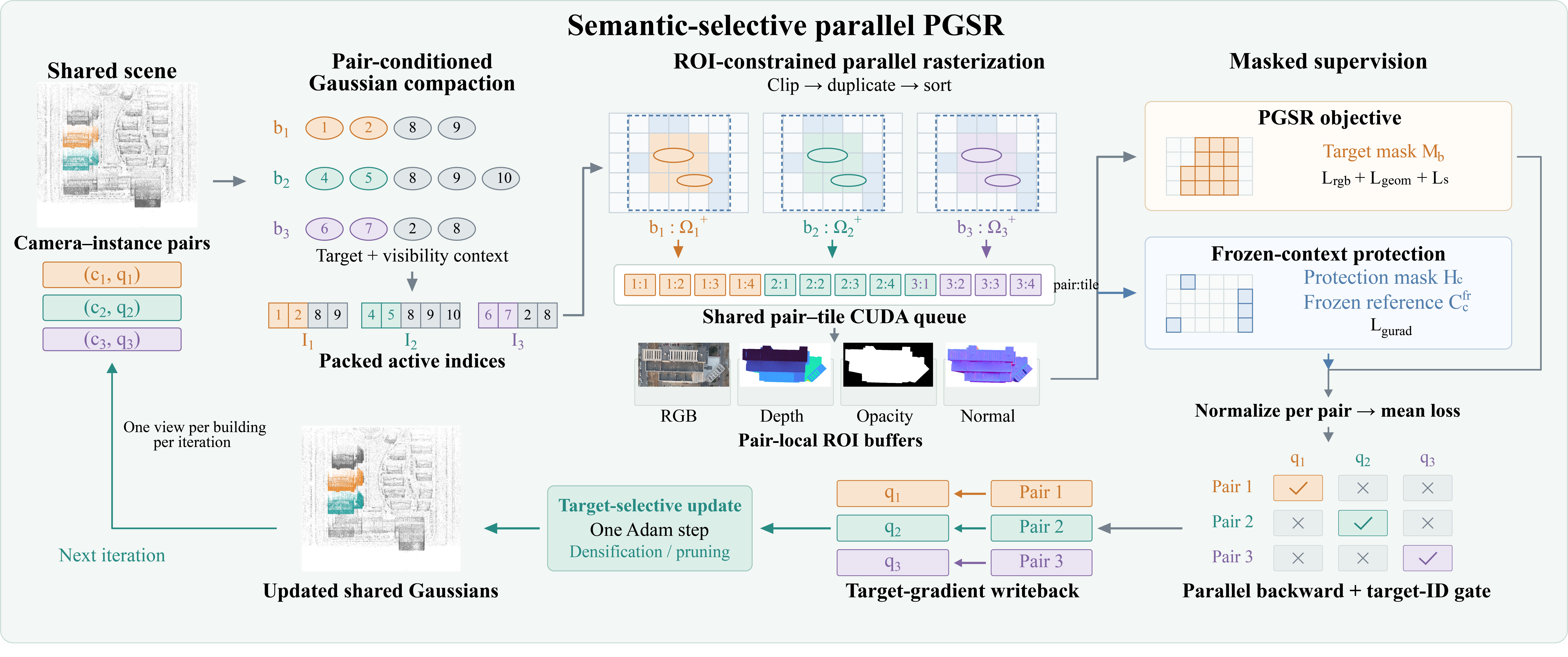}
\caption{Workflow of semantic-selective parallel PGSR. Each iteration selects one valid camera observation for each sampled building. Pair-conditioned compaction packs the active Gaussian indices, and ROI-constrained rasterization processes the resulting pair--tile tasks concurrently to produce pair-local rendering buffers. Target-mask supervision and frozen-context protection guide selective updates through parallel backpropagation and target-gradient writeback. Orange, teal, and purple indicate distinct target instances, while gray denotes contextual Gaussians within each pair. Gaussian numbers identify shared scene indices, and queue entries use the format pair:tile. The return arrow indicates the next optimization iteration. Loss weights are omitted for clarity.}
\label{fig:batched_pgsr_pipeline}
\end{figure*}

Building on the camera--instance observation graph, which provides valid multi-view observations, masks, and ROIs for the target buildings, SemLite-GS refines multiple building instances concurrently within a shared Gaussian representation. Figure~\ref{fig:batched_pgsr_pipeline} illustrates the optimization loop, comprising three stages. Pair-conditioned Gaussian compaction first constructs a compact active set containing the target and contextual Gaussians required for each sampled pair. ROI-constrained parallel rasterization then organizes the Gaussian--tile entries into a shared pair--tile work queue, enabling concurrent rendering across buildings and cameras while restricting tile construction, sorting, and pixel evaluation to the corresponding ROIs. Finally, masked supervision combines the PGSR photometric and geometric terms over verified target observations with frozen-context protection at reliable context pixels near building boundaries. Parallel backpropagation computes pair-local gradients, and only target gradients are written back to the corresponding scene-level indices for selective parameter updates. The updated shared representation is used in the next iteration, enabling parallel training while preserving contextual visibility.

\subsubsection{Pair-Conditioned Gaussian Compaction}
\label{sec:pair_gaussian_compaction}

At each iteration, $B$ distinct building instances are sampled, and one valid camera observation is selected for each instance, forming
\begin{equation}
\mathcal{B}
=
\left\{
(c_b,q_b,M_b,\Omega_b)
\right\}_{b=1}^{B}
\subseteq
\mathcal{E}_{\mathcal{Q}},
\qquad
q_b\neq q_{b'} \;\; \text{for } b\neq b',
\label{eq:camera_instance_batch}
\end{equation}
where $c_b$ and $q_b$ denote the camera and target-building identity of pair $b$, respectively, and $M_b=M_{c_b,q_b}$ and $\Omega_b=\Omega_{c_b,q_b}$ denote its verified mask and ROI. Each building contributes at most one camera--instance pair per iteration, ensuring that the target Gaussian sets of different pairs are disjoint. Sampling is balanced across buildings and their valid cameras over successive iterations. Valid neighboring observations of each reference view provide the correspondences required by multi-view terms of PGSR.

Conventional full-scene Gaussian rasterization projects the entire Gaussian scene for every rendering, incurring the projection and memory overhead associated with full-scene optimization. Therefore, SemLite-GS performs pair-conditioned compaction for all sampled pairs, constructing an active Gaussian domain for each pair:

\begin{equation}
\mathcal{A}_b
=
\left\{i\mid\ell_i=q_b\right\}
\cup
\mathcal{C}_b,
\label{eq:pair_active_gaussians}
\end{equation}

where $\{i\mid\ell_i=q_b\}$ contains the target Gaussians of building $q_b$, and $\mathcal{C}_b$ contains the contextual Gaussians whose projected support may affect depth-ordered compositing within $\Omega_b$, or within the guarded ROI $\Omega_b^{+}$ defined below when protection is enabled. The latter includes foreground terrain, vegetation, and neighboring structures that may occlude the target. Target and contextual Gaussians jointly participate in forward rendering to preserve visibility, but each pair contributes parameter updates only to its target Gaussians. Active sets may overlap through contextual Gaussians; a Gaussian used as context in one pair may still be updated as a target in another pair.

The active Gaussians of each pair are compacted into a pair-local index list and packed into a shared buffer for GPU processing. Let $r_b(i)$ denote the compact index of scene Gaussian $i$ in pair $b$. This representation allows pairs with different active-set sizes to be processed concurrently while preserving the shared scene-level Gaussian parameters.

\subsubsection{ROI-Constrained Parallel Rasterization}
\label{sec:roi_batched_rasterization}

Pair-conditioned compaction restricts the three-dimensional primitive domain, while the ROI confines the corresponding image-space computation. For each camera $c$, a binary protection mask $H_c$ and a frozen-context reference image $\mathbf{C}^{\mathrm{fr}}_c$ are precomputed by rendering the frozen contextual Gaussians. The mask $H_c$ identifies reliable context pixels near target building boundaries, including eligible pixels immediately inside the target mask and in the adjacent background. These pixels are selected using tests based on frozen opacity, consistency with the input image, and trainable opacity and depth near the inner boundary. For pair $b$, the guarded ROI $\Omega_b^{+}$ comprises $\Omega_b$ and the outward annular context region identified by $H_{c_b}$, which provides the neighborhood required for depth-derived normal supervision. 

During preprocessing, the active Gaussians are projected in parallel and clipped to their corresponding guarded ROIs before tile duplication. The Gaussian--tile entries of pair $b$ are defined as
\begin{equation}
\mathcal{K}_b
=
\left\{
(i,t)
\;\middle|\;
i\in\mathcal{A}_b,\;
t\in
\mathcal{T}_{c_b}(g_i)
\cap
\mathcal{T}(\Omega_b^{+})
\right\},
\label{eq:roi_tile_entries}
\end{equation}
where $\mathcal{T}_{c_b}(g_i)$ denotes the tiles intersected by the projected footprint of $g_i$, and $\mathcal{T}(\Omega_b^{+})$ denotes the tiles overlapping the guarded ROI. Each entry stores the pair-local Gaussian index $r_b(i)$ and its camera-space depth $z_{c_bi}$ for subsequent parameter lookup and depth ordering.

Applying the ROI constraint before tile duplication ensures that only Gaussians whose projected footprints overlap $\Omega_b^{+}$ generate tile entries. The entries from all sampled pairs, $\bigcup_{b=1}^{B}\mathcal{K}_b$, are jointly sorted by pair index, tile index, and depth, after which the nonempty pair--tile ranges form a shared CUDA work queue. Gaussians outside the guarded ROI therefore incur no sorting, rasterization, or gradient-computation cost.

Each nonempty pair--tile range is processed by one CUDA thread block using its pair-specific camera parameters, compact active Gaussian set, and local output buffers. Only pixels within $\Omega_b^{+}$ are evaluated by PGSR Gaussian projection and front-to-back compositing; plane-distance and normal maps are converted into depth using Eq.~\ref{eq:prelim_pgsr_depth}. When frozen-context protection is enabled, the protected pixels in $H_{c_b}$ are rendered together with the target region under the same depth ordering. Color,plane-distance, depth, opacity, and normal attributes are written to pair-local ROI buffers, preserving contextual visibility while avoiding full-resolution rendering and output allocation.

\subsubsection{Masked Supervision and Parallel Selective Optimization}
\label{sec:masked_selective_backward}

Rendering and parameter updates involve different Gaussian sets. Target and contextual Gaussians are jointly composited during the forward pass to preserve visibility, whereas only the target Gaussians receive parameter updates. The PGSR photometric and single-view geometric terms are evaluated within the verified target masks, and its multi-view terms use valid neighboring observations, while the additional protection term is evaluated over the precomputed protection masks. For the $B$ camera--instance pairs sampled at an iteration, the masked batch objective is
\begin{equation}
\mathcal{L}_{\mathcal{B}}
=
\frac{1}{B}\sum_{b=1}^{B}
\left[
\mathcal{L}_{\mathrm{rgb}}^{M_b}
+\lambda_1\mathcal{L}_{s}^{q_b}
+\mathcal{L}_{\mathrm{geom}}^{b}
+\lambda_g\mathcal{L}_{\mathrm{guard}}^{H_{c_b}}
\right],
\label{eq:masked_batched_loss}
\end{equation}
Here, $\mathcal{L}_{\mathrm{rgb}}^{M_b}$ and $\mathcal{L}_{\mathrm{svgeom}}^{M_b}$ are evaluated over valid target pixels in $M_b\cap\Omega_b$, while $\mathcal{L}_{s}^{q_b}$ regularizes the Gaussians belonging to target building $q_b$. The set $V_b$ contains valid target-region correspondences between the reference view and its neighboring views; the multi-view terms are evaluated over these correspondences. Each term is normalized over its respective valid pixels, correspondences, or target Gaussians before the pair losses are averaged. The final term, $\mathcal{L}_{\mathrm{guard}}^{H_{c_b}}$, is the frozen-context protection loss. It constrains the rendering at reliable context pixels near target boundaries, with $\lambda_g$ controlling its weight.

For pair $b$, let $\mathcal{H}_b=\{\mathbf{p}\in\Omega_b^{+}\mid H_{c_b}(\mathbf{p})=1\}$ denote the protected pixels. The protection loss measures the channel-averaged RGB difference between the current target-plus-context rendering and the frozen-context reference:
\begin{equation}
\mathcal{L}^{H_{c_b}}_{\mathrm{guard}}
=
\frac{1}{3\max(1,|\mathcal{H}_b|)}
\sum_{\mathbf{p}\in\mathcal{H}_b}
\left\|
\widehat{\mathbf{C}}_{b}(\mathbf{p})
-
\mathbf{C}^{\mathrm{fr}}_{c_b}(\mathbf{p})
\right\|_1
\label{eq:frozen_context_guard_loss}
\end{equation}

Pairs with no protected pixels make no contribution to $\mathcal{L}_{\mathrm{guard}}$. Normalization by the number of protected pixels prevents pairs with larger protected regions from dominating the batch objective. The loss therefore penalizes deviations from the frozen-context rendering at reliable context pixels near target boundaries.

During backpropagation, each pair accumulates gradients independently, and contributions associated with the same scene Gaussian are subsequently reduced across pairs. The optimizer receives
\begin{equation}
\mathbf{g}^{\mathrm{upd}}_i
=
\sum_{\substack{b\\i\in\mathcal{A}_b}}
\mathbb{I}[\ell_i=q_b]\,
\mathbf{g}_{b,r_b(i)},
\label{eq:target_gated_gradient}
\end{equation}
where $\mathbf{g}_{b,r_b(i)}$ is the pair-local gradient of Gaussian $g_i$. The indicator restricts parameter updates, densification, and pruning to the target Gaussians of the corresponding building, while contextual Gaussians remain fixed. The optimized representation of building $q$ is denoted by $\mathcal{G}_q^\star=\{g_i^\star\mid\ell_i=q\}$, whose rendered plane-derived depth, opacity, and normal observations are used for the plane-constrained lightweight mesh reconstruction in Sec.~\ref{sec:mesh_generation}.

\subsection{Plane-Constrained Lightweight Mesh Reconstruction}
\label{sec:mesh_generation}

\begin{figure*}[!tp]
    \centering
    \includegraphics[width=\textwidth]{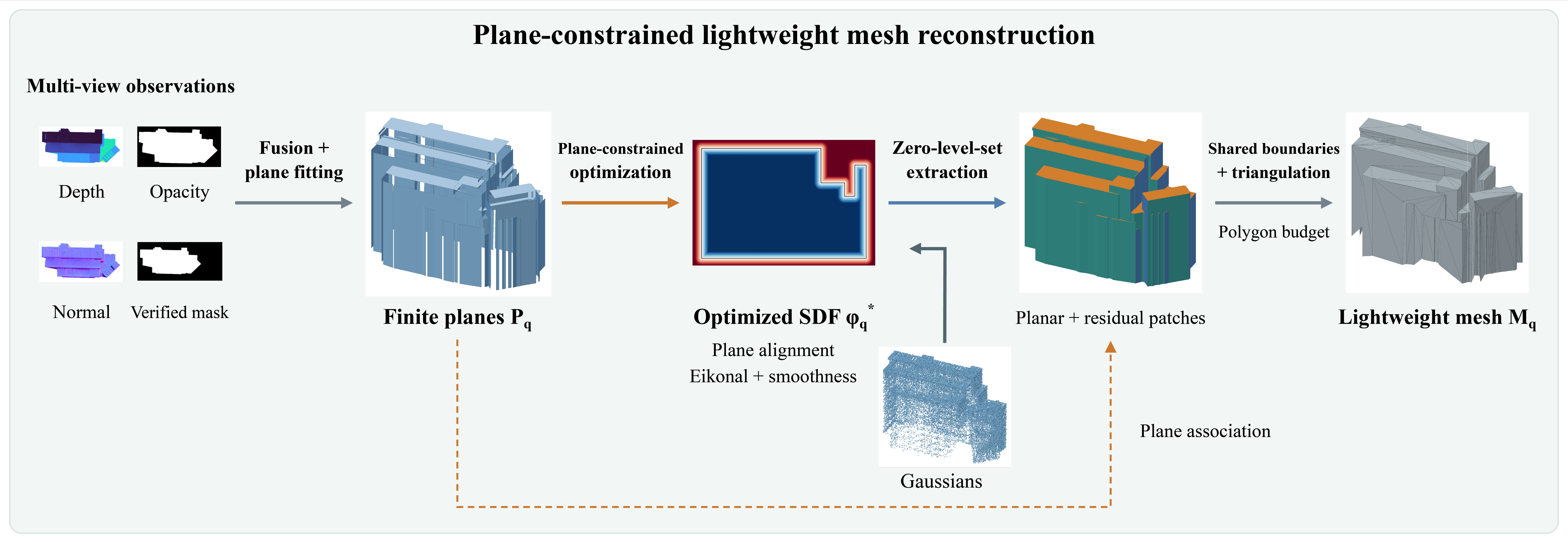}
    \caption{
    Pipeline of plane-constrained lightweight mesh reconstruction.
    Multi-view depth, opacity, and normal observations within verified
    target masks are fused to estimate finite architectural planes.
    These planes regularize a Gaussian-initialized SDF and guide the
    consolidation of planar surface regions during extraction.
    Planar and residual patches are triangulated with shared boundaries
    under a prescribed polygon budget.
    The dashed arrow indicates the reuse of finite plane supports
    for plane association.
    }
    \label{fig:plane_mesh_pipeline}
\end{figure*}

Although $\mathcal{G}_q^\star$ captures detailed surface geometry, its Gaussian primitives do not explicitly encode the architectural structure needed to allocate a limited polygon budget. As illustrated in Figure~\ref{fig:plane_mesh_pipeline}, SemLite-GS reconstructs lightweight building meshes through three stages. Multi-view geometric observations are first fused to estimate finite architectural planes. These planes constrain an instance-level SDF initialized from the optimized Gaussian geometry. The refined SDF and finite plane supports then jointly guide surface extraction and triangulation, producing compact planar regions while preserving nonplanar surfaces and shared boundaries. This process directly generates lightweight meshes without constructing dense intermediate meshes for generic post-hoc decimation.

\subsubsection{Multi-View Architectural Plane Extraction}
\label{sec:plane_extraction}

Plane estimates derived from a single rendered view are susceptible to occlusion, grazing viewing angles, and local reconstruction noise. SemLite-GS therefore combines geometric observations from the valid camera--instance pairs of each building. For each retained camera--instance pair $(c,q,M_{c,q},\Omega_{c,q})\in\mathcal{E}_{\mathcal{Q}}$,  $\mathcal{G}_q^\star$ is rendered to obtain the depth map $D_{c,q}^\star$, opacity map $A_{c,q}^\star$, and normal map $\mathbf{N}_{c,q}^\star$.

Within the target masks, pixels with valid depth, sufficient opacity, consistent rendered and depth-derived normals, and suitable viewing geometry are back-projected to world coordinates. Each retained pixel contributes a position $\mathbf{x}_m$, normal $\mathbf{n}_m$, and confidence $w_m$, where the confidence combines opacity, normal consistency, viewing angle, and distance from the mask boundary. Observations from all valid views are accumulated in a confidence-weighted voxel grid and grouped by spatial connectivity and normal similarity. For each candidate group $\mathcal{S}_{q,k}$, a plane is estimated by jointly fitting its positions and normals:

\begin{equation}
\begin{aligned}
(\mathbf{n}_{q,k},d_{q,k})
={}&
\underset{\|\mathbf{n}\|_2=1,\,d}{\operatorname{argmin}}
\sum_{m\in\mathcal{S}_{q,k}} w_m
\Bigl[
\psi\!\left(\mathbf{n}^{\mathsf T}\mathbf{x}_m+d\right)\\
&\quad
+\lambda_o
\left(1-\left|\mathbf{n}^{\mathsf T}\mathbf{n}_m\right|\right)
\Bigr]
\end{aligned}
\label{eq:weighted_plane_fitting}
\end{equation}

where $\psi(\cdot)$ is a point-to-plane penalty and $\lambda_o$ balances positional fitting and normal alignment. The robust penalty limits the influence of residual outliers, while the normal-alignment term stabilizes the estimated plane orientation.

The fitted planes are subsequently screened according to their spatial extent, fitting quality, and multi-view support. For each retained plane, the inlier observations are projected into local plane coordinates to define a compact finite footprint $\mathcal{R}_{q,k}$. The resulting finite-plane set for building $q$ is

\begin{equation}
\mathcal{P}_q
=
\left\{
(\mathbf{n}_{q,k},
d_{q,k},
\mathcal{R}_{q,k},
\omega_{q,k})
\right\}_{k=1}^{K_q},
\label{eq:architectural_plane_set}
\end{equation}

where $K_q$ is the number of retained planes and $\omega_{q,k}$ denotes the confidence of plane $k$. Each footprint restricts its associated plane to the observed roof or facade region. The resulting set provides finite architectural priors for the subsequent SDF refinement and planar region consolidation.

\subsubsection{Plane-Constrained SDF Optimization}
\label{sec:sdf_optimization}

The finite-plane set $\mathcal{P}_q$ provides local architectural constraints but does not by itself define a complete building surface. SemLite-GS therefore constructs an instance-level voxel grid around building $q$ and initializes an outside-positive SDF $\phi_q^0$ by fusing surface positions and normals derived from $\mathcal{G}_q^\star$. This initialization provides a continuous initial building envelope without requiring an intermediate polygonal mesh.

Each plane in $\mathcal{P}_q$ is rasterized only within its finite footprint $\mathcal{R}_{q,k}$, where it provides point-to-plane, normal, and confidence constraints. The plane influence is tapered near footprint boundaries, and overlapping constraints near architectural intersections are combined according to their confidence and spatial proximity. Voxels without reliable plane constraints remain governed by the Gaussian. The SDF is refined through a spatially localized correction:

\begin{equation}
\phi_q(\mathbf{x})
=
\phi_q^0(\mathbf{x})
+
\eta_q(\mathbf{x})\delta_q(\mathbf{x}),
\qquad
\label{eq:sdf_correction}
\end{equation}

where $\eta_q(\mathbf{x})\in[0,1]$ is the spatial influence map induced by the finite plane footprints, and $\delta_q(\mathbf{x})$ is the learned correction field. This parameterization confines plane-induced deformation to regions supported by reliable architectural observations.

The correction field is optimized to align the locally adjusted surface with the plane priors while preserving a regular signed-distance field:
\begin{equation}
\mathcal{L}_{\mathrm{SDF}}
=
\mathcal{L}_{\mathrm{plane}}
+
\lambda_e\mathcal{L}_{\mathrm{eik}}
+
\mathcal{L}_{\mathrm{smooth}},
\label{eq:sdf_objective}
\end{equation}

where $\mathcal{L}_{\mathrm{plane}}$ aligns the zero level set and its gradient with the finite plane priors, weighted by $\omega_{q,k}$ within $\mathcal{R}_{q,k}$. The Eikonal term $\mathcal{L}_{\mathrm{eik}}$ encourages the field to retain its signed-distance property, while $\mathcal{L}_{\mathrm{smooth}}$ suppresses local variation in the correction field. Together, these terms regularize planar regions while maintaining coherent geometry near plane junctions and in regions without reliable plane support. The optimized field $\phi_q^\star$ is passed together with $\mathcal{P}_q$ to the mesh extraction stage.

\subsubsection{Direct Lightweight Mesh Extraction}
\label{sec:lightweight_mesh_generation}

The optimized SDF $\phi_q^\star$ provides a continuous building surface, while $\mathcal{P}_q$ identifies regions that can be represented by compact planar geometry. Grid cells containing a sign change of $\phi_q^\star$ are identified, and their zero-crossing positions and normals are interpolated from the field. A sample is assigned to plane $k$ only when it lies within $\mathcal{R}_{q,k}$ and satisfies the corresponding distance and normal-consistency criteria.

For each sign-changing voxel cell, the interpolated zero crossings define a local zero-level-set surface fragment. Surface fragments in adjacent cells are stitched across their shared grid faces, producing an initial connectivity graph of the extracted surface. Samples associated with the same plane are consolidated before triangulation, and their boundaries are simplified in local plane coordinates before projection back to three-dimensional space. Samples without reliable plane support are retained as adaptive residual regions. Planar and residual regions share the same boundary vertices, preserving the connectivity of roof ridges, facade junctions, corners, and plane intersections.

The resulting planar and residual patches are triangulated using the shared surface connectivity to obtain the lightweight mesh $\mathcal{M}_q$. A prescribed polygon budget governs boundary simplification within planar regions and sample retention within residual regions before face construction. Degenerate elements and insignificant disconnected components are subsequently removed. Repeating this extraction procedure over all target building instances yields $\{\mathcal{M}_q\mid q\in\mathcal{Q}\}$ directly from the optimized SDF, thereby avoiding the reconstruction and subsequent generic decimation of dense intermediate meshes.

\FloatBarrier
\section{Experiments}
\label{sec:experiments}

\subsection{Experimental Setup}
\label{sec:experimental_setup}

\subsubsection{Datasets}
\label{sec:datasets}

The evaluation encompasses six large-scale UAV scenes drawn from GauScene \citep{xiong2024gauuscene}, Skylume \citep{li2026skylume}, and Tianjin\_Xiqing \citep{lin2025true}. LiDAR reference point clouds are available for the five GauScene and Skylume scenes, enabling quantitative assessment; Tianjin\_Xiqing is included only in the qualitative comparisons. Table~\ref{tab:datasets} summarizes the image count, resolution, and number of target building instances for each scene. All analyses are based on the registered multi-view UAV images and associated camera parameters provided by the datasets.

\begin{table*}[!t]
\centering
\caption{Overview of the selected scenes. Resolution denotes the dimensions (width $\times$ height in pixels) of the highest-pixel-count image available for each scene. Target instances indicates the number of building instances in each scene.}
\label{tab:datasets}
\small
\setlength{\tabcolsep}{6pt}
\begin{tabular}{cccccc}
\toprule
Dataset & Scene & \# Images & Resolution & Target instances & LiDAR \\
\midrule
GauScene & CUHK\_LOWER & 670   & $5472 \times 3648$ & 18 & Yes \\
         & CUHK\_UPPER & 715   & $5472 \times 3648$ & 22 & Yes \\
         & SZIT        & 1,215 & $5472 \times 3648$ & 26 & Yes \\
\midrule
Skylume & building     & 3,646 & $1563 \times 1039$ & 24 & Yes \\
        & tec\_school  & 2,394 & $6260 \times 4165$ & 11 & Yes \\
\midrule
Tianjin\_Xiqing      & Tianjin\_Xiqing           & 305   & $3942 \times 2930$ & 34 & No \\
\bottomrule
\end{tabular}
\end{table*}

\subsubsection{Implementation Details}
\label{sec:implementation_details}

All experiments were conducted on a workstation equipped with a single NVIDIA RTX 4090 GPU. For image-based pipelines, input images from GauScene and Skylume were downsampled by a factor of eight, whereas
those from Tianjin\_Xiqing were downsampled by a factor of four. SemLite-GS performs shared scene-level initialization followed by semantic-selective parallel PGSR training of the selected buildings.

\subsubsection{Evaluation Metrics}
\label{sec:evaluation_metrics}

Following GS2Poly \citep{liu2025gs2poly}, lightweight building reconstruction is evaluated in terms of geometric accuracy and mesh compactness. Geometric accuracy is measured using Chamfer distance (CD), while mesh compactness is characterized by the numbers of vertices and faces.

\textbf{Geometric accuracy.}
For all LiDAR-based CD evaluations, the reference points for each building were manually extracted from the scene-level point cloud and downsampled to a spatial resolution of 0.1~m. Let $\mathcal{P}$ denote points sampled uniformly with respect to surface area from the reconstructed mesh, and let $\mathcal{P}_{\mathrm{ref}}$ denote the corresponding processed LiDAR points in the same coordinate system. The point-to-set distance is defined as
\begin{equation}
d(\mathbf{x},\mathcal{S})
=
\min_{\mathbf{y}\in\mathcal{S}}
\|\mathbf{x}-\mathbf{y}\|_2,
\end{equation}
where $\|\cdot\|_2$ denotes the Euclidean norm.

CD measures the mean bidirectional discrepancy between the reconstructed and reference point sets:
\begin{equation}
\operatorname{CD}(\mathcal{P},\mathcal{P}_{\mathrm{ref}})
=
\frac{1}{2}
\left[
\frac{1}{|\mathcal{P}|}
\sum_{\mathbf{p}\in\mathcal{P}}
d(\mathbf{p},\mathcal{P}_{\mathrm{ref}})
+
\frac{1}{|\mathcal{P}_{\mathrm{ref}}|}
\sum_{\mathbf{r}\in\mathcal{P}_{\mathrm{ref}}}
d(\mathbf{r},\mathcal{P})
\right].
\label{eq:chamfer_distance}
\end{equation}
Here $|\cdot|$ denotes set cardinality. As CD uses non-squared Euclidean distances, it has the same physical unit as the scene coordinates; lower values indicate better geometric agreement.

\textbf{Lightweight modeling quality.}
Mesh compactness is quantified by the vertex count ($N_v$) and polygonal face count ($N_f$). These metrics are evaluated alongside CD to assess the trade-off between geometric fidelity and compactness.

\subsection{Results and Analysis}
\label{sec:comparison}

The evaluation comprises three parts. First, SemLite-GS is compared with CityGS-X, CityGaussianV2, PG-SAG, and Metashape to assess computational efficiency and lightweight mesh quality. Second, the proposed mesh reconstruction module is compared with alternative lightweight reconstruction methods using fixed instance-level Gaussian representations optimized by SemLite-GS as their geometric input. Third, ablation studies examine the efficiency and cross-view consistency of instance-mask propagation, the contribution of the frozen-context protection loss, and the effect of parallel camera--instance pair training.

\subsubsection{Comparison with Large-Scale Reconstruction Pipelines}
\label{sec:large_scale_comparison}

To the best of our knowledge, existing methods do not provide an automated solution for the lightweight reconstruction of selected buildings from large-scale UAV scenes. GS2Poly \citep{liu2025gs2poly} is closely related to this study because it generates compact building meshes from 3DGS-derived representations. However, it assumes that a representation of the target building has already been isolated and does not automatically localize building instances within a large scene. The absence of a public GS2Poly implementation at the time of this study precluded a reproducible quantitative comparison. Accordingly, CityGS-X \citep{gao2025citygsx}, CityGaussianV2 \citep{liu2025citygaussianv2}, and PG-SAG \citep{wang2025pgsag} were selected as representative state-of-the-art open-source baselines for large-scale Gaussian-based reconstruction. Metashape \citep{metashape} was included as a photogrammetric baseline for comparison with a conventional dense-scene reconstruction workflow.

The comparison spans six scenes: five from GauScene and Skylume and the Tianjin\_Xiqing scene. Metashape reconstructs dense full-scene meshes from the input images. Target buildings are manually extracted from these meshes and evaluated both before simplification and, as Metashape+QEM, after QEM simplification. CityGS-X and CityGaussianV2 reconstruct dense full-scene meshes under their default configurations; target buildings are then manually extracted and simplified using QEM. PG-SAG retains its original mask-generation and building-reconstruction pipeline, followed by QEM simplification. SemLite-GS directly reconstructs lightweight meshes for automatically identified building instances. To isolate the contribution of mesh generation, SemLite-GS (TSDF)+QEM fuses depth maps rendered from the Gaussian representation optimized by SemLite-GS into a dense mesh and subsequently applies QEM. To assess mask transferability, PG-SAG (mask)+QEM replaces PG-SAG's original masks with those generated by SemLite-GS while keeping its remaining reconstruction and QEM settings fixed. The comparison examines instance-level semantic selectivity, lightweight mesh quality, and computational runtime.

\paragraph{\textbf{Instance-Level Semantic Selectivity.}}

\begin{figure*}[!tp]
\centering
\includegraphics[width=\linewidth]{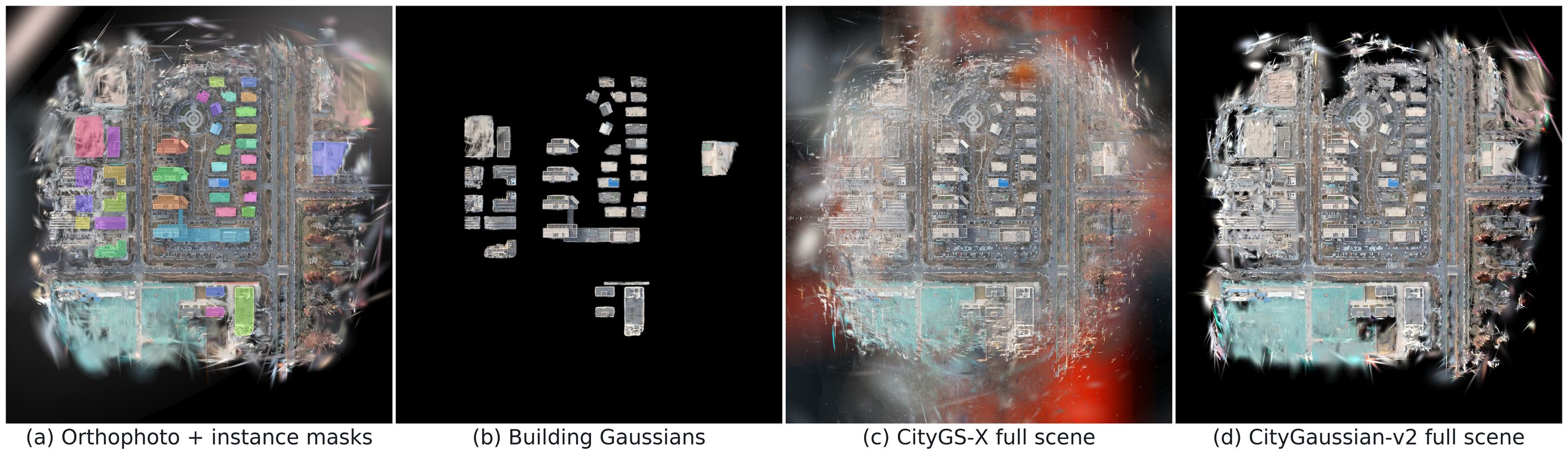}
\caption{Instance-level semantic selectivity on the Tianjin\_Xiqing scene. (a) Orthophoto overlaid with automatically generated building-instance masks. (b) Gaussian rendering of the corresponding building instances after selective PGSR training by SemLite-GS. (c) Full-scene Gaussian reconstruction by CityGS-X. (d) Full-scene Gaussian reconstruction by CityGaussianV2. The panels are shown from aligned overhead viewpoints.}
\label{fig:spatial_selectivity}
\end{figure*}

Instance-level semantic selectivity is evaluated by examining whether target buildings are identified as distinct reconstruction units before Gaussian training and mesh generation. Figure~\ref{fig:spatial_selectivity} illustrates this capability on the Tianjin\_Xiqing scene. Panels (a) and (b) show the orthographic building-instance masks and the corresponding Gaussian representations after selective PGSR training. Panels (c) and (d) show the full-scene reconstructions produced by CityGS-X and CityGaussianV2.

The correspondence between the masks in Fig.~\ref{fig:spatial_selectivity}(a) and the Gaussian representations in Fig.~\ref{fig:spatial_selectivity}(b) shows that SemLite-GS retains building identities when transferring the orthographic instance map to three-dimensional reconstruction. These identities define target-specific training and mesh-generation domains, while surrounding Gaussians remain available for visibility reasoning. CityGS-X and CityGaussianV2 instead reconstruct the complete scene, requiring target buildings to be identified and extracted afterward. The runtime implications of this difference are quantified in Table~\ref{tab:large_scale_efficiency}.

\paragraph{\textbf{Lightweight Reconstruction Quality.}}

This evaluation assesses the geometric accuracy and architectural fidelity of lightweight building meshes reconstructed from large-scale scenes. Qualitative comparisons cover all six scenes, while quantitative evaluation uses the five GauScene and Skylume scenes with LiDAR references. For each evaluated building, the corresponding LiDAR points were manually isolated from the scene-level point cloud and downsampled to a spatial resolution of 0.1~m before metric computation.

\begin{figure*}[!t]
\centering
\includegraphics[
  width=\textwidth,
  height=0.90\textheight,
  keepaspectratio
]{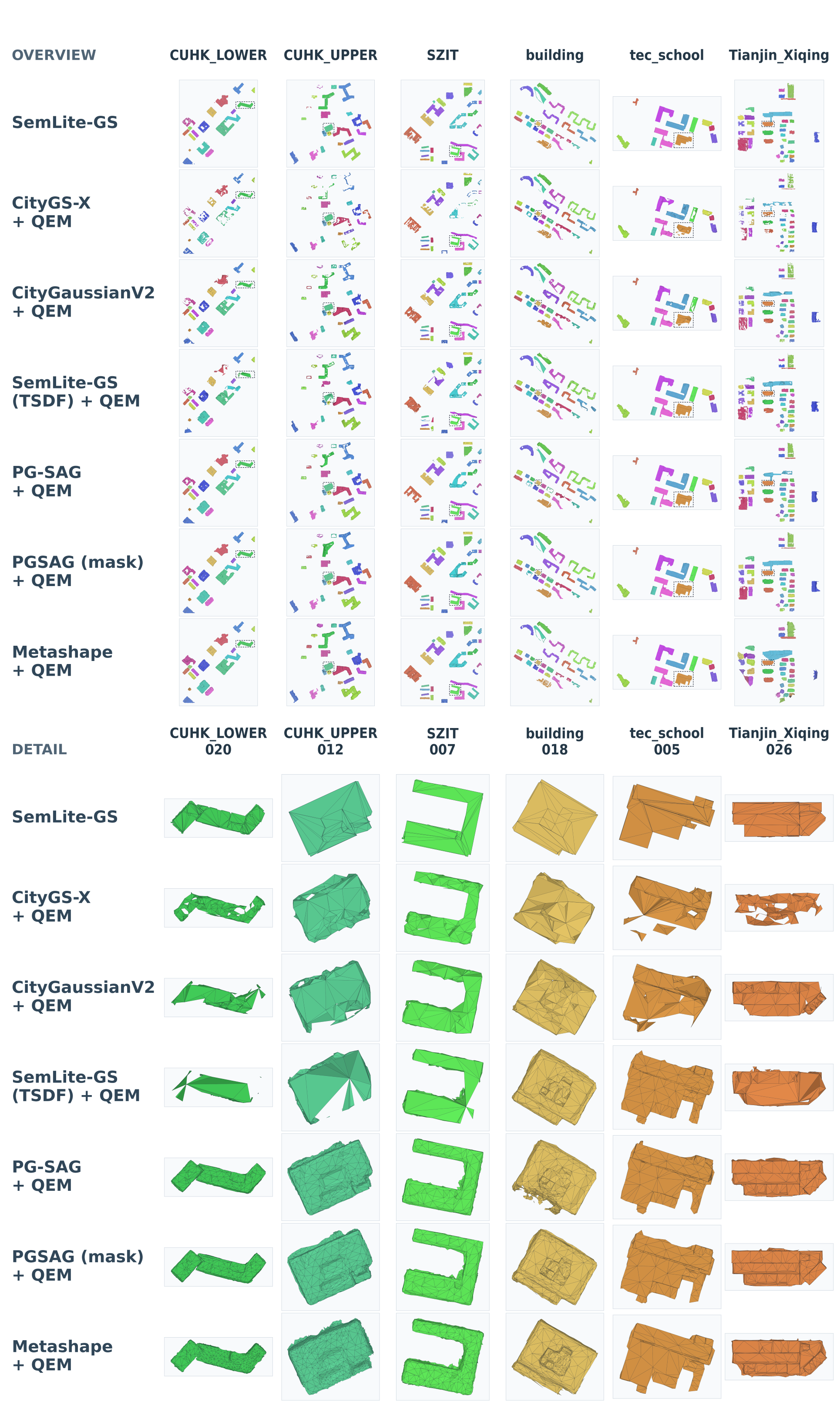}
\caption{Qualitative comparison of lightweight building meshes produced
by seven pipelines across six scenes. Columns show CUHK\_LOWER, CUHK\_UPPER, SZIT, building, tec\_school, and Tianjin\_Xiqing. The upper rows present scene overviews; dashed boxes identify the buildings shown from oblique viewpoints in the lower rows. Corresponding building instances use consistent colors across methods.}
\label{fig:large_scale_mesh_comparison}
\end{figure*}

Figure~\ref{fig:large_scale_mesh_comparison} compares seven lightweight mesh configurations across six scenes. In the detailed oblique views, SemLite-GS represents broad roof and facade planes with relatively sparse triangulation while retaining distinct ridges, corners, and roof--facade junctions. Metashape+QEM preserves the overall building forms, but several planar regions exhibit more irregular triangulation and less clearly delineated architectural edges. CityGS-X+QEM and CityGaussianV2+QEM show locally fragmented surfaces and irregular boundaries. Relative to SemLite-GS (TSDF)+QEM, which shares the same optimized Gaussian representation, SemLite-GS produces more continuous planar regions and better-defined structural transitions. PG-SAG+QEM captures the general building forms, while substituting SemLite-GS masks improves the apparent completeness of selected buildings. These examples provide qualitative context for the CD and mesh-complexity results reported below.

Table~\ref{tab:quality_large_scale} reports per-building CD and mesh complexity, averaged within each scene and across all 101 evaluated buildings. To control mesh complexity, QEM-based outputs were limited to 1,000 faces per building in four scenes and 400 faces in tec\_school. These budgets reflect the adaptive output of SemLite-GS, which averages 821--998 faces in the former scenes and 362 faces in tec\_school.

SemLite-GS achieves the lowest CD in all five evaluated scenes. Its overall mean CD of 0.914~m is 15.1\% lower than that of the next best configuration, PG-SAG (mask)+QEM, at 1.077~m. SemLite-GS reduces the overall mean CD by 18.2\% relative to Metashape+QEM (1.117~m). Compared with SemLite-GS (TSDF)+QEM, which uses the same optimized Gaussian representation, the proposed mesh reconstruction reduces the overall CD from 1.272 to 0.914~m, a 28.1\% reduction. This comparison isolates the contribution of mesh generation from that of Gaussian optimization. The accuracy gain is accompanied by an average of 880 faces per building, compared with 934--935 recorded faces for the QEM-based configurations, although the vertex count of SemLite-GS is not the lowest in every comparison. Vertex count alone is an insufficient measure of lightweight mesh quality. Some QEM outputs retain oppositely oriented duplicate triangles, which count toward the prescribed face budget while sharing existing vertices and adding no distinct surface geometry. SemLite-GS contains no such duplicates and preserves vertices along architectural boundaries, providing a closer correspondence between mesh complexity and the represented building structure.

Replacing PG-SAG's original building masks with those generated by SemLite-GS reduces CD in every evaluated scene. With the remaining PG-SAG reconstruction and QEM settings held fixed, the substitution lowers the overall mean CD from 1.360 to 1.077~m, a 20.8\% reduction. This result indicates that the proposed masks can improve geometric accuracy when incorporated into a separate mask-based reconstruction pipeline.

\begin{table*}[!t]
\centering
\caption{Scene-level comparison of geometric accuracy and mesh complexity for lightweight building meshes across five scenes with LiDAR references. Metashape+QEM uses building meshes extracted from the Metashape scene reconstruction and simplified using QEM. PG-SAG (mask)+QEM replaces PG-SAG's original building masks with those generated by SemLite-GS. QEM-based meshes are simplified to 1,000 faces per building in four scenes and 400 faces in tec\_school, whereas SemLite-GS determines the face count adaptively for each building. CD, vertex count, and face count are computed per building and averaged within each scene. The Overall row reports instance-weighted means over all 101 evaluated buildings.}
\label{tab:quality_large_scale}
\footnotesize
\setlength{\tabcolsep}{4pt}
\renewcommand{\arraystretch}{1.05}
\begin{tabular}{ccclccc}
\toprule
Dataset & Scene & Valid instances & Pipeline
& CD (m)$\downarrow$ & Avg. vertices$\downarrow$ & Avg. faces$\downarrow$ \\
\midrule
\multirow[c]{21}{*}{GauScene}
& \multirow[c]{7}{*}{CUHK\_LOWER} & \multirow[c]{7}{*}{18}
& CityGS-X + QEM           & 2.305 & 282 & 1000 \\
& & & CityGaussianV2 + QEM    & 1.559 & 384 & 1000 \\
& & & SemLite-GS (TSDF) + QEM & 1.528 & 407 & 1000 \\
& & & PG-SAG + QEM            & 1.373 & 563 & 1000 \\
& & & PG-SAG (mask) + QEM     & 1.031 & 578 & 1000 \\
& & & Metashape + QEM         & 1.154 & 582 & 1000 \\
& & & SemLite-GS              & \textbf{1.026} & 546 & 998 \\
\cmidrule(lr){2-7}
& \multirow[c]{7}{*}{CUHK\_UPPER} & \multirow[c]{7}{*}{22}
& CityGS-X + QEM           & 2.990 & 318 & 1000 \\
& & & CityGaussianV2 + QEM    & 1.370 & 402 & 1000 \\
& & & SemLite-GS (TSDF) + QEM & 1.261 & 341 & 1000 \\
& & & PG-SAG + QEM            & 1.130 & 547 & 1000 \\
& & & PG-SAG (mask) + QEM     & 1.069 & 545 & 1000 \\
& & & Metashape + QEM         & 1.170 & 562 & 1000 \\
& & & SemLite-GS              & \textbf{0.948} & 542 & 984 \\
\cmidrule(lr){2-7}
& \multirow[c]{7}{*}{SZIT} & \multirow[c]{7}{*}{26}
& CityGS-X + QEM           & 2.367 & 290 & 1000 \\
& & & CityGaussianV2 + QEM    & 1.560 & 414 & 1000 \\
& & & SemLite-GS (TSDF) + QEM & 1.550 & 399 & 1000 \\
& & & PG-SAG + QEM            & 1.524 & 565 & 1000 \\
& & & PG-SAG (mask) + QEM     & 1.191 & 569 & 1000 \\
& & & Metashape + QEM         & 1.590 & 600 & 1000 \\
& & & SemLite-GS              & \textbf{1.131} & 543 & 986 \\
\midrule
\multirow[c]{14}{*}{Skylume}
& \multirow[c]{7}{*}{building} & \multirow[c]{7}{*}{24}
& CityGS-X + QEM           & 1.915 & 368 & 1000 \\
& & & CityGaussianV2 + QEM    & 1.677 & 462 & 1000 \\
& & & SemLite-GS (TSDF) + QEM & 1.063 & 642 & 1000 \\
& & & PG-SAG + QEM            & 1.613 & 658 & 1000 \\
& & & PG-SAG (mask) + QEM     & 1.198 & 647 & 1000 \\
& & & Metashape + QEM         & 0.710 & 698 & 1000 \\
& & & SemLite-GS              & \textbf{0.702} & 450 & 821 \\
\cmidrule(lr){2-7}
& \multirow[c]{7}{*}{tec\_school} & \multirow[c]{7}{*}{11}
& CityGS-X + QEM           & 1.573 & 140 & 400 \\
& & & CityGaussianV2 + QEM    & 0.963 & 166 & 400 \\
& & & SemLite-GS (TSDF) + QEM & 0.677 & 223 & 400 \\
& & & PG-SAG + QEM            & 0.857 & 233 & 400 \\
& & & PG-SAG (mask) + QEM     & 0.632 & 239 & 400 \\
& & & Metashape + QEM         & 0.723 & 250 & 400 \\
& & & SemLite-GS              & \textbf{0.616} & 203 & 362 \\
\midrule
\multirow[c]{7}{*}{Overall}
& \multirow[c]{7}{*}{Five scenes} & \multirow[c]{7}{*}{101}
& CityGS-X + QEM           & 2.298 & 297 & 935 \\
& & & CityGaussianV2 + QEM    & 1.481 & 391 & 935 \\
& & & SemLite-GS (TSDF) + QEM & 1.272 & 426 & 935 \\
& & & PG-SAG + QEM            & 1.360 & 547 & 935 \\
& & & PG-SAG (mask) + QEM     & 1.077 & 548 & 935 \\
& & & Metashape + QEM         & 1.117 & 574 & 935 \\
& & & SemLite-GS              & \textbf{0.914} & 484 & 880 \\
\bottomrule
\end{tabular}
\end{table*}

\paragraph{\textbf{Surface Reconstruction Quality.}}

To assess the capability of SemLite-GS for detailed building surface reconstruction, dense meshes are generated for the Gaussian-based methods by applying the same TSDF fusion procedure to multi-view depth maps rendered from their optimized Gaussian representations. The Metashape dense mesh is obtained through its photogrammetric reconstruction pipeline. All meshes are evaluated before QEM simplification on the same 101 building instances with LiDAR references. CD is computed for each building following the protocol described above and averaged within each scene.

As shown in Table~\ref{tab:quality_surface_reconstruction}, SemLite-GS achieves the lowest CD in all five evaluated scenes, with an overall mean of 0.737~m, 13.0\% below PG-SAG (mask), the next best method at 0.847~m. Metashape achieves an overall mean CD of 1.082~m; SemLite-GS reduces this error by 31.9\% and achieves a lower CD in every scene. Replacing PG-SAG's original masks with those generated by SemLite-GS reduces its overall mean CD from 1.110 to 0.847~m, a 23.7\% reduction. These results demonstrate that SemLite-GS also achieves a substantial geometric-accuracy advantage in dense building surface reconstruction.

\begin{table*}[!tp]
\centering
\caption{Geometric accuracy of dense building surface meshes before QEM simplification. Gaussian-based methods extract meshes by TSDF fusion of rendered multi-view depth maps, whereas Metashape produces meshes through its photogrammetric reconstruction pipeline. PG-SAG (mask) denotes the mask-substitution variant described above. CD is computed per building and averaged within each scene; the overall row averages all 101 evaluated buildings.}
\label{tab:quality_surface_reconstruction}
\small
\setlength{\tabcolsep}{3pt}
\renewcommand{\arraystretch}{1.05}
\begin{tabular*}{\textwidth}{@{\extracolsep{\fill}}llccccccc@{}}
\toprule
& & & \multicolumn{6}{c}{CD (m)$\downarrow$} \\
\cmidrule(lr){4-9}
Dataset & Scene & Valid instances
& SemLite-GS & PG-SAG & PG-SAG (mask)
& CityGaussianV2 & CityGS-X & Metashape \\
\midrule
\multirow[c]{3}{*}{GauScene}
& CUHK\_LOWER & 18 & \textbf{0.833} & 1.306 & 0.952 & 1.214 & 1.455 & 1.140 \\
& CUHK\_UPPER & 22 & \textbf{0.807} & 0.972 & 0.902 & 1.033 & 1.904 & 1.165 \\
& SZIT         & 26 & \textbf{0.972} & 1.344 & 1.022 & 1.241 & 1.633 & 1.574 \\
\midrule
\multirow[c]{2}{*}{Skylume}
& building    & 24 & \textbf{0.481} & 1.043 & 0.693 & 0.705 & 1.148 & 0.611 \\
& tec\_school & 11 & \textbf{0.445} & 0.657 & 0.484 & 0.601 & 0.984 & 0.681 \\
\midrule
Overall & Five scenes & 101
& \textbf{0.737} & 1.110 & 0.847 & 0.994 & 1.474 & 1.082 \\
\bottomrule
\end{tabular*}
\end{table*}

\paragraph{\textbf{Computational Efficiency.}}

Computational efficiency is assessed by the cumulative wall-clock time required to reconstruct lightweight meshes of target buildings across the five scenes. For the Gaussian-based pipelines, runtime is divided into preprocessing, training, and mesh reconstruction. For CityGS-X and CityGaussianV2, preprocessing comprises depth prediction, depth calibration, and input validation, followed by full-scene Gaussian optimization. The CityGaussianV2 runtime also includes scene partitioning and checkpoint merging. For both methods, mesh reconstruction comprises dense mesh extraction and QEM simplification of manually identified target buildings; manual interaction time is excluded. For a consistent input basis, Metashape received the same registered UAV images and camera poses as the 3DGS-based methods. The measured wall-clock time spans input import, dense mesh reconstruction, instance extraction, and QEM simplification, excluding manual interaction. For PG-SAG, preprocessing includes building-mask generation and preparation of instance-specific inputs, training comprises Gaussian optimization for the building instances, and mesh reconstruction includes dense mesh extraction followed by QEM simplification. For SemLite-GS, preprocessing comprises FastGS initialization, orthographic rendering and segmentation, and multi-view instance propagation. Its subsequent stages consist of semantic-selective parallel PGSR training and plane-constrained lightweight mesh reconstruction.

\begin{table*}[!tp]
\centering
\caption{Cumulative wall-clock runtime for lightweight building-mesh reconstruction across five scenes. Stage-level times are reported for the 3DGS-based pipelines. For Metashape+QEM, only the total software execution time is reported, covering registered-image and camera-pose import, dense-scene mesh reconstruction, instance extraction, and QEM simplification. Manual target selection is excluded.}
\label{tab:large_scale_efficiency}
\small
\setlength{\tabcolsep}{6pt}
\begin{tabular}{lcccc}
\toprule
Pipeline & Preprocessing (h)$\downarrow$ & Training (h)$\downarrow$
& Mesh reconstruction (h)$\downarrow$ & Total time (h)$\downarrow$ \\
\midrule
CityGS-X + QEM        & 0.40 & 12.19 & 1.49 & 14.08 \\
CityGaussianV2 + QEM & 0.43 & 72.38 & 1.06 & 73.87 \\
PG-SAG + QEM          & 1.07 & 9.29  & 0.64 & 11.00 \\
Metashape + QEM       & --   & --    & 3.57   & 3.57 \\
SemLite-GS            & 0.64 & \textbf{2.31} & \textbf{0.53} & \textbf{3.48} \\
\bottomrule
\end{tabular}
\end{table*}

Table~\ref{tab:large_scale_efficiency} shows that SemLite-GS requires 3.48~h across the five scenes, compared with 14.08~h for CityGS-X+QEM, 73.87~h for CityGaussianV2+QEM, and 11.00~h for PG-SAG+QEM. These correspond to runtime reductions of 75.3\%, 95.3\%, and 68.4\%, respectively. SemLite-GS also has the shortest Gaussian training and mesh reconstruction times among these pipelines. Its amortized end-to-end runtime is approximately 124~s per building. Metashape+QEM requires 3.57~h, close to the 3.48~h of SemLite-GS. As a mature commercial photogrammetry system with extensive performance optimizations relative to standard open-source workflows such as COLMAP \citep{schonberger2016sfm}, Metashape provides a demanding efficiency baseline; SemLite-GS nevertheless retains a modest runtime advantage.

The runtime differences among the Gaussian-based pipelines reflect their computational scope and data reuse. CityGS-X and CityGaussianV2 optimize and mesh complete scenes before extracting target buildings. PG-SAG reconstructs building instances in parallel, but the evaluated implementation prepares inputs and renders views separately for each camera--instance task, while building masks do not constrain every optimization operation. SemLite-GS instead shares image data and coordinates ROI-constrained rasterization through a pair--tile queue, restricting parameter updates to target Gaussians while retaining contextual Gaussians for visibility. These design choices reduce redundant computation and contribute to its shorter end-to-end runtime.

\FloatBarrier
\subsubsection{Comparison with Lightweight Mesh Reconstruction Methods}
\label{sec:lightweight_comparison}

Complementing the comparison with large-scale reconstruction pipelines, this experiment evaluates the proposed mesh-generation module against lightweight reconstruction methods applied to individual buildings. The instance-level Gaussian representations optimized by SemLite-GS are held fixed to isolate differences arising from mesh generation. Geometric observations derived from these representations are converted into the inputs required by PolyFit \citep{nan2017polyfit}, Kinetic Shape Reconstruction (KSR) \citep{bauchet2020kinetic}, Compod \citep{sulzer2024compod}, and the proposed module.

\begin{figure*}[!t]
\centering
\includegraphics[
  width=\textwidth,
  height=0.72\textheight,
  keepaspectratio
]{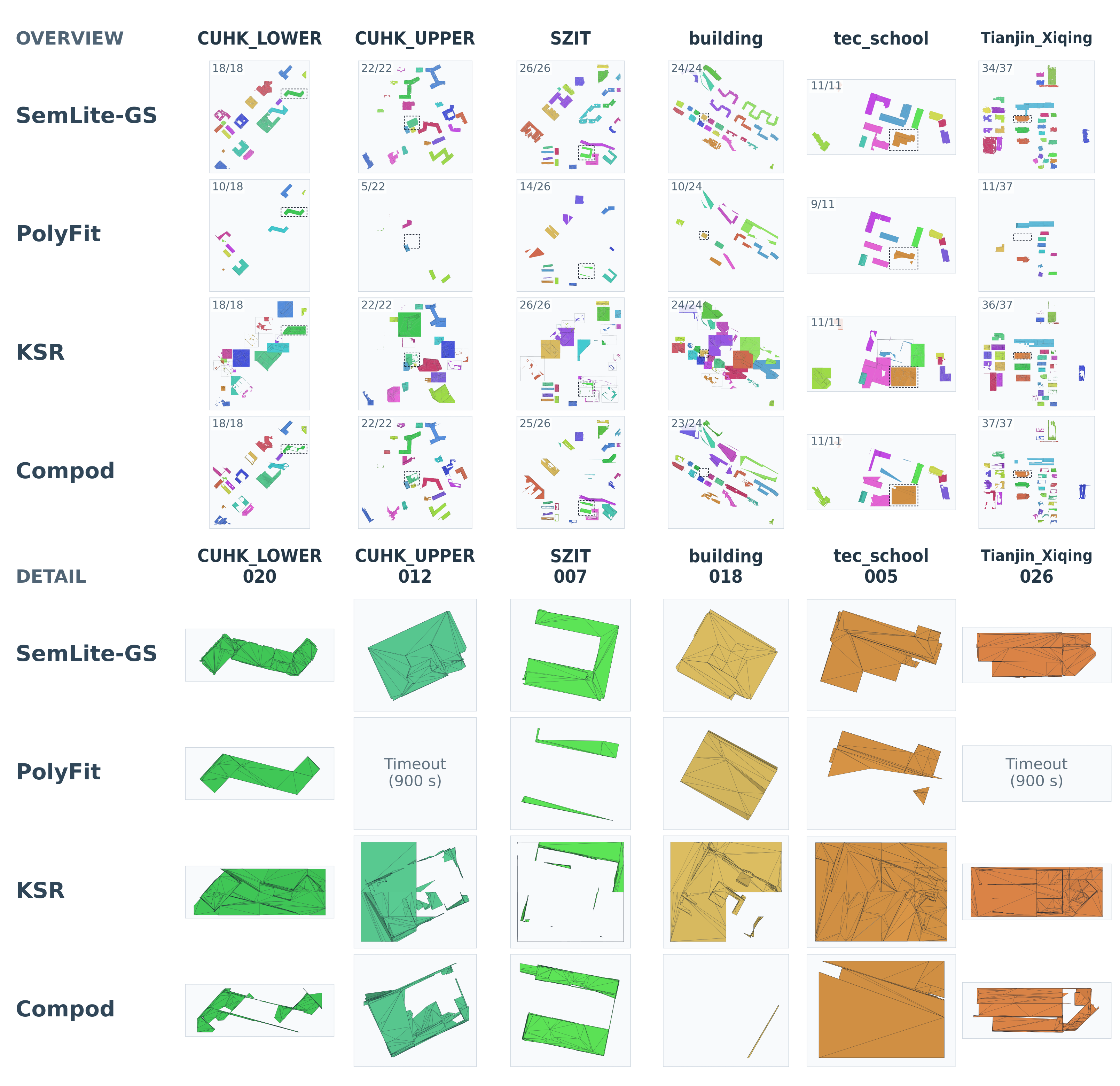}
\caption{Qualitative comparison of lightweight building meshes reconstructed by SemLite-GS, PolyFit, KSR, and Compod across six scenes. Columns correspond to CUHK\_LOWER, CUHK\_UPPER, SZIT, building, tec\_school, and Tianjin\_Xiqing. The upper four rows show scene overviews; the fractions indicate successfully reconstructed buildings relative to the target count. The lower four rows show enlarged views of buildings marked by dashed boxes. ``Timeout (900~s)'' indicates that PolyFit did not produce a mesh for the selected building within the allotted time. Colors denote corresponding building instances across methods. Tianjin\_Xiqing is included for qualitative assessment only because LiDAR reference data are unavailable.}
\label{fig:quality_shared_visual}
\end{figure*}

Table~\ref{tab:quality_shared_frontend} reports geometric accuracy and mesh complexity. All metrics are computed independently for each building and subsequently within each scene. Because PolyFit and Compod do not reconstruct every target, their aggregate CD values are computed over method-specific successful subsets; reconstruction coverage is therefore considered alongside geometric accuracy.

\begin{table*}[!t]
\centering
\caption{Comparison of lightweight mesh reconstruction methods across five scenes. Instance-level Gaussian representations optimized by SemLite-GS provide a common geometric basis. PolyFit, KSR, and Compod receive identical Gaussian-center point clouds, whereas SemLite-GS generates meshes directly from the corresponding optimized Gaussian representations. Superscripts report successful reconstructions relative to the number of target instances. Metrics are averaged over successful instances within each scene, and overall values pool successful instances across scenes. The evaluated building subsets therefore differ by method.}
\label{tab:quality_shared_frontend}
\small
\setlength{\tabcolsep}{4pt}
\renewcommand{\arraystretch}{1.05}
\begin{tabular}{ccclccc}
\toprule
Dataset & Scene & Valid instances & Method
& CD (m)$\downarrow$
& Avg. vertices$\downarrow$
& Avg. faces$\downarrow$ \\
\midrule
\multirow[c]{12}{*}{GauScene}
& \multirow[c]{4}{*}{CUHK\_LOWER}
& \multirow[c]{4}{*}{18}
& PolyFit$^{10/18}$       & 3.380 & 204 & 208 \\
& & & KSR                & 5.729 & 441 & 792 \\
& & & Compod             & 2.670 & 414 & 860 \\
& & & SemLite-GS         & \textbf{1.026} & 546 & 998 \\
\cmidrule(lr){2-7}
& \multirow[c]{4}{*}{CUHK\_UPPER}
& \multirow[c]{4}{*}{22}
& PolyFit$^{5/22}$        & 3.906 & 115 & 101 \\
& & & KSR                & 4.215 & 414 & 772 \\
& & & Compod             & 1.765 & 412 & 842 \\
& & & SemLite-GS         & \textbf{0.948} & 542 & 984 \\
\cmidrule(lr){2-7}
& \multirow[c]{4}{*}{SZIT}
& \multirow[c]{4}{*}{26}
& PolyFit$^{14/26}$       & 4.118 & 148 & 161 \\
& & & KSR                & 7.701 & 447 & 757 \\
& & & Compod$^{25/26}$   & 2.886 & 434 & 920 \\
& & & SemLite-GS         & \textbf{1.131} & 543 & 986 \\
\midrule
\multirow[c]{8}{*}{Skylume}
& \multirow[c]{4}{*}{building}
& \multirow[c]{4}{*}{24}
& PolyFit$^{10/24}$       & 4.497 & 198 & 167 \\
& & & KSR                & 3.039 & 530 & 779 \\
& & & Compod$^{23/24}$   & 2.862 & 462 & 677 \\
& & & SemLite-GS         & \textbf{0.702} & 450 & 821 \\
\cmidrule(lr){2-7}
& \multirow[c]{4}{*}{tec\_school}
& \multirow[c]{4}{*}{11}
& PolyFit$^{9/11}$        & 3.188 & 192 & 175 \\
& & & KSR                & 5.878 & 185 & 318 \\
& & & Compod             & 3.849 & 173 & 322 \\
& & & SemLite-GS         & \textbf{0.616} & 203 & 362 \\
\midrule
\multirow[c]{4}{*}{Overall}
& \multirow[c]{4}{*}{Five scenes}
& \multirow[c]{4}{*}{101}
& PolyFit$^{48/101}$      & 3.847 & 175 & 168 \\
& & & KSR                & 5.284 & 430 & 724 \\
& & & Compod$^{99/101}$  & 2.699 & 403 & 769 \\
& & & SemLite-GS         & \textbf{0.914} & 484 & 880 \\
\bottomrule
\end{tabular}
\end{table*}

Figure~\ref{fig:quality_shared_visual} compares reconstruction coverage and local mesh geometry across six scenes. In the five scenes with LiDAR references, SemLite-GS reconstructs all 101 target buildings, whereas PolyFit succeeds on 48; the CUHK\_UPPER and Tianjin\_Xiqing examples also show PolyFit reaching the 900~s time limit. KSR reconstructs all 101 evaluated buildings and Compod reconstructs 99, but the enlarged views reveal extraneous geometry and irregular surfaces in KSR results, as well as missing surfaces or disrupted boundaries in several Compod results. SemLite-GS produces more continuous roof and facade surfaces with clearly defined architectural transitions in the examples shown. These observations complement the reconstruction-success counts and CD measurements in Table~\ref{tab:quality_shared_frontend}.

\subsubsection{Ablation Studies}
\label{sec:ablation}

Ablation studies on the building and tec\_school scenes from Skylume assess orthographically anchored mask propagation, frozen-context protection, and parallel camera--instance pair training. The evaluated instances and controlled settings are specified for each experiment.

\paragraph{\textbf{Cross-View Consistency and Efficiency of Orthographically Anchored Mask Propagation.}}

Orthographically anchored mask propagation was evaluated against independent SAM3 \citep {carion2026sam} segmentation of each oblique view. Both approaches were evaluated on the same registered images and target buildings with identical text prompt (\texttt{buildings}). Baseline predictions were matched to fixed projected instance labels according to their overlap, assigning target identities without altering the predicted mask boundaries. No manual corrections were applied. Three representative buildings were selected to assess cross-view consistency under substantial variation in viewing direction, image scale, and visible extent.

\begin{figure*}[!tp]
\centering
\includegraphics[width=\linewidth]{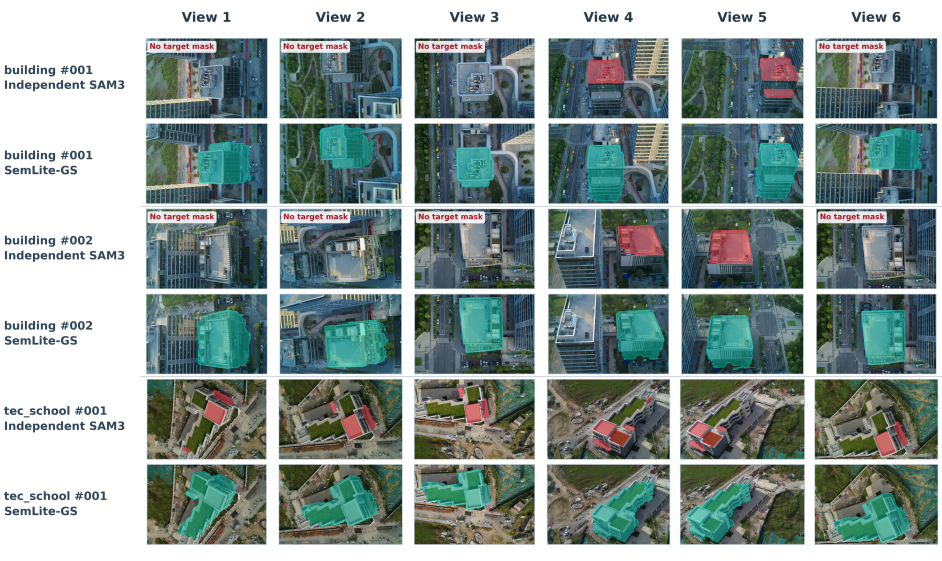}
\caption{Cross-view comparison of instance masks for three representative buildings from the building and tec\_school scenes. Columns correspond to six oblique views. Within each pair of rows, the upper row shows independent per-view SAM3 masks in red, and the lower row shows the masks propagated by SemLite-GS in cyan. ``No target mask'' indicates that no independent SAM3 prediction was associated with the target building in the corresponding view. Target association is determined using fixed projected instance labels, while the displayed mask boundaries are obtained directly from the independent SAM3 predictions.}
\label{fig:ablation_mask_consistency}
\end{figure*}

Figure~\ref{fig:ablation_mask_consistency} compares the resulting masks across six oblique viewpoints. For instances \#001 and \#002 in the building scene, no independent SAM3 prediction could be associated with the target in four of the six views, although the buildings remained visible. For instance \#001 in tec\_school, the associated predictions repeatedly covered isolated roof sections rather than the full visible building region. Orthographically anchored propagation maintained the building identities across views and produced more complete and coherent masks. These examples illustrate its ability to mitigate missed associations and mask fragmentation under changes in viewpoint and scale. 

Table~\ref{tab:ablation_mask_efficiency} reports mask-generation times measured on the same NVIDIA RTX 4090 GPU. Independent SAM3 processed every oblique image sequentially using the same prompt with SemLite-GS; its reported time includes model loading, initial warm-up, image loading, and inference. The SemLite-GS time includes orthographic rendering, SAM3 segmentation of the orthophoto, mask construction, and visibility-aware propagation to the registered oblique views. SemLite-GS completed mask generation in 298.0~s for the 3,646-image building scene and 148.1~s for the 2,394-image tec\_school scene, compared with 1,031.3~s and 677.5~s for independent SAM3, respectively, demonstrating efficient semantic propagation across image collections containing thousands of oblique views.

\begin{table}[!tp]
\centering
\caption{Runtime of multi-view instance-mask generation on an NVIDIA RTX 4090 GPU. Independent SAM3 processes all oblique images sequentially, whereas SemLite-GS segments overlapping DOM tiles and propagates the resulting instance labels to the registered views through parallel visibility-aware rendering.}
\label{tab:ablation_mask_efficiency}
\small
\setlength{\tabcolsep}{5pt}
\begin{tabular}{cccc}
\toprule
Scene & Images & Independent SAM3 (s)$\downarrow$ & SemLite-GS (s)$\downarrow$ \\
\midrule
building & 3,646 & 1,031.3 & 298.0 \\
tec\_school & 2,394 & 677.5 & 148.1 \\
\bottomrule
\end{tabular}
\end{table}

\paragraph{\textbf{Effect of the Frozen-Context Protection Loss.}}

The contribution of the frozen-context protection loss $\mathcal{L}_{\mathrm{guard}}$ (Eq.~\ref{eq:frozen_context_guard_loss}) is assessed by comparing PGSR training with and without this term ($\lambda_g=0$). The two configurations share the same camera--instance pairs, target masks, fixed contextual Gaussians, and optimization schedules.

\begin{figure*}[!tp]
\centering
\includegraphics[width=\linewidth]{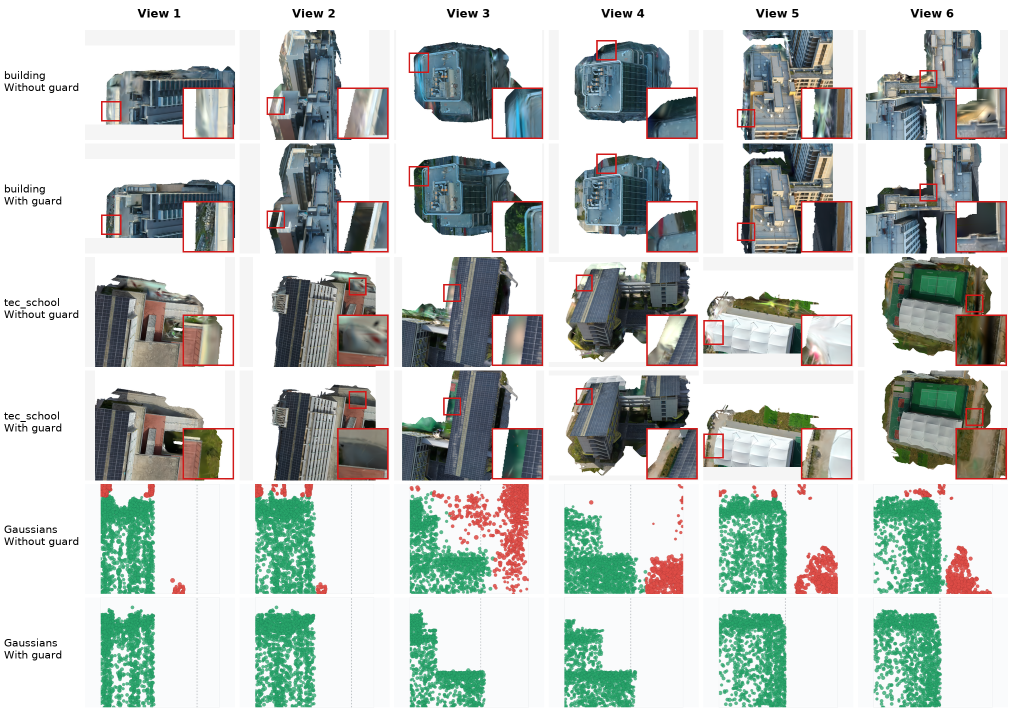}
\caption{Visual effect of the frozen-context protection loss $\mathcal{L}_{\mathrm{guard}}$. The upper four rows show renderings of six representative buildings from the building and tec\_school scenes, each observed from two viewpoints. Within each scene, results without the loss precede those with the loss. Red boxes and enlarged insets identify boundary leakage and spurious connections. The bottom two rows show lateral views of the corresponding local Gaussian distributions without and with $\mathcal{L}_{\mathrm{guard}}$, respectively. Green denotes building-associated Gaussians, and red marks spurious Gaussians near building boundaries.}
\label{fig:ablation_guard_loss}
\end{figure*}

Figure~\ref{fig:ablation_guard_loss} compares the resulting renderings and local Gaussian distributions. Boundary errors in the instance masks may assign pixels from vegetation or adjacent structures to a target building. Without $\mathcal{L}_{\mathrm{guard}}$, these misassigned observations can drive target Gaussians to fit surrounding content, producing boundary leakage and artificial connections in the rendered views. The lateral views reveal Gaussian primitives extending beyond the building boundaries, indicating that these artifacts originate in the three-dimensional representation. Enabling $\mathcal{L}_{\mathrm{guard}}$ suppresses the extraneous geometry and yields clearer separation between the target buildings and their surroundings.

Table~\ref{tab:ablation_guard_loss} quantifies the effect of $\mathcal{L}_{\mathrm{guard}}$ on dense meshes extracted by TSDF fusion and lightweight meshes produced by plane-constrained reconstruction. Enabling the loss reduces CD for both mesh types in each scene. Across all 35 buildings, mean CD decreases from 0.526 to 0.470~m for TSDF meshes and from 0.703 to 0.675~m for lightweight meshes, corresponding to reductions of 10.6\% and 4.0\%, respectively. The improvement across both mesh-generation procedures, together with the visual evidence in Fig.~\ref{fig:ablation_guard_loss}, indicates that frozen-context protection limits boundary contamination during Gaussian training and improves the resulting building geometry.

\begin{table}[!t]
\centering
\caption{Effect of the frozen-context protection loss $\mathcal{L}_{\mathrm{guard}}$ on the geometric accuracy of dense TSDF and lightweight building meshes. CD values are averaged over the building instances in each scene; the overall row reports the mean across all 35 instances.}
\label{tab:ablation_guard_loss}
\footnotesize
\setlength{\tabcolsep}{3pt}
\renewcommand{\arraystretch}{1.05}
\begin{tabular*}{\columnwidth}{@{\extracolsep{\fill}}cccc@{}}
\toprule
& & \multicolumn{2}{c}{CD (m)$\downarrow$} \\
\cmidrule(lr){3-4}
Scene & $\mathcal{L}_{\mathrm{guard}}$ & TSDF mesh & Lightweight mesh \\
\midrule
\multirow{2}{*}{building (24)}
& Without & 0.556 & 0.727 \\
& With    & \textbf{0.481} & \textbf{0.702} \\
\midrule
\multirow{2}{*}{tec\_school (11)}
& Without & 0.460 & 0.652 \\
& With    & \textbf{0.445} & \textbf{0.616} \\
\midrule
\multirow{2}{*}{Overall (35)}
& Without & 0.526 & 0.703 \\
& With    & \textbf{0.470} & \textbf{0.675} \\
\bottomrule
\end{tabular*}
\end{table}

\paragraph{\textbf{Efficiency of Parallel Camera--Instance Pair Training.}}

To quantify the effects of camera--instance pair parallelism on training time and peak GPU memory, the number of pairs processed in parallel was varied over $B\in\{1,2,4,8\}$. Each single-process configuration ran on an NVIDIA RTX 4090 GPU, with the training process bound to one physical core of an Intel Xeon Gold 6133 CPU (2.50~GHz). The same 24 instances in building and eight instances in tec\_school were included at every setting. Both instance counts are divisible by each tested value of $B$, ensuring fully populated groups. Per-instance initialization, eligible reference-view pools, and the training objective were held fixed. At each training step, one eligible view was sampled for each active instance, and the resulting camera--instance pairs were processed in parallel.

Table~\ref{tab:ablation_parallel_efficiency} reports cumulative training-loop time and peak GPU memory. Increasing $B$ reduced training time in both scenes. At $B=8$, training was $2.23\times$ faster than at $B=1$ on building and $2.22\times$ faster on tec\_school. When multiple pairs use the same camera view, parallel processing permits reuse of image data and shared rendering infrastructure instead of loading and rendering that view independently for each instance. This limits duplicated buffers and allows GPU memory to be allocated more effectively to trainable Gaussian parameters. Despite the eightfold increase in pairs processed in parallel, peak memory increased by $2.55\times$ on building and $3.08\times$ on tec\_school. The sublinear speedups are consistent with pair-specific computations and contention for shared GPU resources. Among the evaluated single-process settings, $B=8$ achieved the shortest training time, accompanied by the highest peak memory.

\begin{table}[!tp]
\centering
\caption{Cumulative training-loop time and peak GPU memory for different numbers of camera--instance pairs processed in parallel. The same building instances are used at every setting; times are summed across all training groups.}
\label{tab:ablation_parallel_efficiency}
\small
\setlength{\tabcolsep}{4pt}
\begin{tabular}{lccc}
\toprule
Scene & Pairs ($B$) & Time (s)$\downarrow$ & Peak memory (GiB)$\downarrow$ \\
\midrule
\multirow{4}{*}{building}
& 1 & 6693.5 & \textbf{8.95} \\
& 2 & 4867.8 & 10.39 \\
& 4 & 3766.2 & 15.58 \\
& 8 & \textbf{2998.0} & 22.86 \\
\midrule
\multirow{4}{*}{tec\_school}
& 1 & 2214.2 & \textbf{5.25} \\
& 2 & 1593.2 & 6.22 \\
& 4 & 1146.3 & 9.98 \\
& 8 & \textbf{999.2} & 16.19 \\
\bottomrule
\end{tabular}
\end{table}

Although camera--instance pairs are processed in parallel on the GPU, the single-process configuration assigns pair preparation and workload submission to one CPU core, potentially constraining training throughput. To assess the benefit of distributing these operations, additional experiments assigned four pairs to each training process (Table~\ref{tab:ablation_multiprocess_efficiency}). On building, two processes reduced training time from 2998.0~s at the single-process $B=8$ setting to 2555.2~s, a further reduction of 14.8\%. Three processes, each handling four pairs, reduced the time to 2404.1~s. Retaining the inputs on the GPU exceeded device memory in the multi-process building configurations, necessitating CPU-resident inputs; training nevertheless remained faster than in the single-process setting. On tec\_school, GPU-resident inputs remained feasible, and two processes reduced training time from 999.2 to 690.7~s, a reduction of 30.9\%. The larger two-process gain observed under GPU-resident input storage highlights the additional throughput available when memory capacity permits. Together, these results show that distributing pair preparation and workload submission across CPU processes further improves the efficiency of SemLite-GS's parallel training strategy.

\begin{table}[!htbp]
\centering
\caption{Training time and peak GPU memory under multi-process pair submission. Configuration gives the number of processes $\times$ pairs per process; parentheses indicate where the inputs reside.}
\label{tab:ablation_multiprocess_efficiency}
\small
\setlength{\tabcolsep}{3pt}
\begin{tabular}{@{}llrr@{}}
\toprule
Scene & Configuration & Time (s)$\downarrow$ & GPU (GiB)$\downarrow$ \\
\midrule
\multirow{2}{*}{building (24)}
& $2\times4$ (CPU) & 2555.2 & 19.41 \\
& $3\times4$ (CPU) & \textbf{2404.1} & 22.91 \\
\midrule
tec\_school (8)
& $2\times4$ (GPU) & \textbf{690.7} & 20.05 \\
\bottomrule
\end{tabular}
\end{table}

\section{Discussion}
\label{sec:discussion}

SemLite-GS depends on the Gaussian representation initialized by FastGS to render the orthophoto from which globally consistent building identities are established. Incomplete geometry or rendering errors at this stage may leave buildings undetected or produce incomplete instance masks. Such errors can propagate to Gaussian labels and camera--instance observations. Because subsequent semantic-selective PGSR training operates only on identified instances, buildings omitted at initialization cannot be recovered downstream. Future work should explore rapid scene-initialization strategies tailored to orthographic rendering that improve building coverage and instance-mask reliability while preserving computational efficiency.

Architectural fidelity also remains difficult to assess from LiDAR point clouds alone. CD quantifies average geometric proximity but does not explicitly measure planar consistency, boundary straightness, ridge localization, or the integrity of wall intersections. Consequently, a mesh may attain a low CD despite fragmented planes or distorted architectural edges. Future benchmarks should incorporate expert-constructed lightweight building meshes as structural references and supplement distance-based evaluation with measures of planarity, edge and intersection-line deviation, topological correctness, and mesh compactness. Such a protocol would provide a more complete assessment of geometric accuracy and architectural structure preservation.

\section{Conclusion}
\label{sec:conclusion}

This study presents SemLite-GS, a novel building-centric Gaussian splatting framework for lightweight building reconstruction from large-scale UAV oblique imagery. SemLite-GS establishes globally consistent building identities through orthographically anchored instance propagation. Guided by these identities, the framework performs semantic-selective parallel PGSR training while preserving contextual visibility and generates lightweight meshes through plane-constrained reconstruction with SDF optimization. Extensive experiments on multiple large-scale UAV datasets demonstrate that the proposed framework improves cross-view instance consistency, substantially reduces measured computational runtime compared to complete-scene 3DGS pipelines, and achieves a favorable balance among geometric fidelity, architectural structure preservation, and mesh compactness under comparable polygon budgets. Ablation results support the contributions of orthographic anchoring, frozen-context protection, and parallel camera--instance pair training. The reported results collectively indicate that incorporating semantic target selection into Gaussian optimization and architectural constraints into mesh generation provides an effective alternative to dense full-scene mesh extraction followed by post-hoc mesh simplification for on-demand lightweight building modeling. In the future, we will investigate rapid scene-initialization strategies for more reliable orthographic instance discovery and develop structural reference meshes and metrics to evaluate the architectural fidelity of lightweight building reconstructions more comprehensively.


\printcredits

\section*{Acknowledgements}
This work was supported by the National Natural Science Foundation of China under Grant 42671570 awarded to Zongqian Zhan, and by the National Natural Science Foundation of China under Grant 42301507 awarded to Xin Wang.
\bibliographystyle{cas-model2-names}

\bibliography{references}



\end{document}